\documentclass[11pt,a4paper]{article}
\usepackage[hyperref]{emnlp2020}
\usepackage{times}
\usepackage{latexsym}

\usepackage{amsfonts}       
\usepackage{amssymb}
\usepackage{amsmath}
\usepackage{todonotes}
\usepackage{float}
\usepackage{array,multirow,graphicx}
\usepackage[export]{adjustbox}
\usepackage{graphicx}
\usepackage{hyperref}
\usepackage{booktabs}

\usepackage{microtype}

\aclfinalcopy 

\title{Pruning Redundant Mappings in Transformer Models \\via Spectral-Normalized Identity Prior}

\author{Zi Lin\thanks{\ \ Work done as part of the Google AI Residency.} \\
  Google Research \\
  \texttt{lzi@google.com} \\\And
  Jeremiah Zhe Liu\thanks{\ \ Work done at Google Research.} \\
  Google Research \& Harvard University \\
  \texttt{jereliu@google.com} \\\And
  Zi Yang \\
  Google Research \\
  \texttt{ziy@google.com} \\\AND
  Nan Hua \\
  Google Research \\
  \texttt{nhua@google.com} \\\And
  Dan Roth \\
  University of Pennsylvania \\
  \texttt{danroth@seas.upenn.edu}
  }

\date{}

\begin{document}
\maketitle
\begin{abstract}
  Traditional (unstructured) pruning methods for a Transformer model focus on regularizing the individual weights by penalizing them toward zero.
  In this work, we explore \textit{spectral-normalized identity priors} (SNIP), a structured pruning approach that penalizes an entire residual module in a Transformer model toward an identity mapping. Our method identifies and discards unimportant non-linear mappings in the residual connections by applying a thresholding operator on the function norm. It is applicable to any structured module, including a single attention head, an entire attention block, or a feed-forward subnetwork. 
  Furthermore, we introduce \textit{spectral normalization} to stabilize the distribution of the post-activation values of the Transformer layers, further improving the pruning effectiveness of the proposed methodology.
  We conduct experiments with BERT  on 5 GLUE benchmark tasks to demonstrate that SNIP achieves effective pruning results while maintaining comparable performance. Specifically, we improve the performance over the state-of-the-art by 0.5 to 1.0\% on average at 50\% compression ratio.
\end{abstract}

\section{Introduction}

Natural Language Processing (NLP) has recently achieved great success by using the Transformer-based pre-trained models \cite{radford2019language, devlin2018bert, yang2019xlnet, clark2020electra}. However, these models often consume considerable storage, memory bandwidth, and computational resource. To reduce the model size and increase the inference throughput, compression techniques such as knowledge distillation \cite{sanh2019distilbert, sun2019patient, tang2019distilling, jiao2019tinybert, sun2020mobilebert} and weight pruning \cite{guo2019reweighted, wang2019structured, gordon2020compressing, sanh2020movement} have recently been developed.

\begin{figure}[t]
    \centering
    \includegraphics[width=7.5cm]{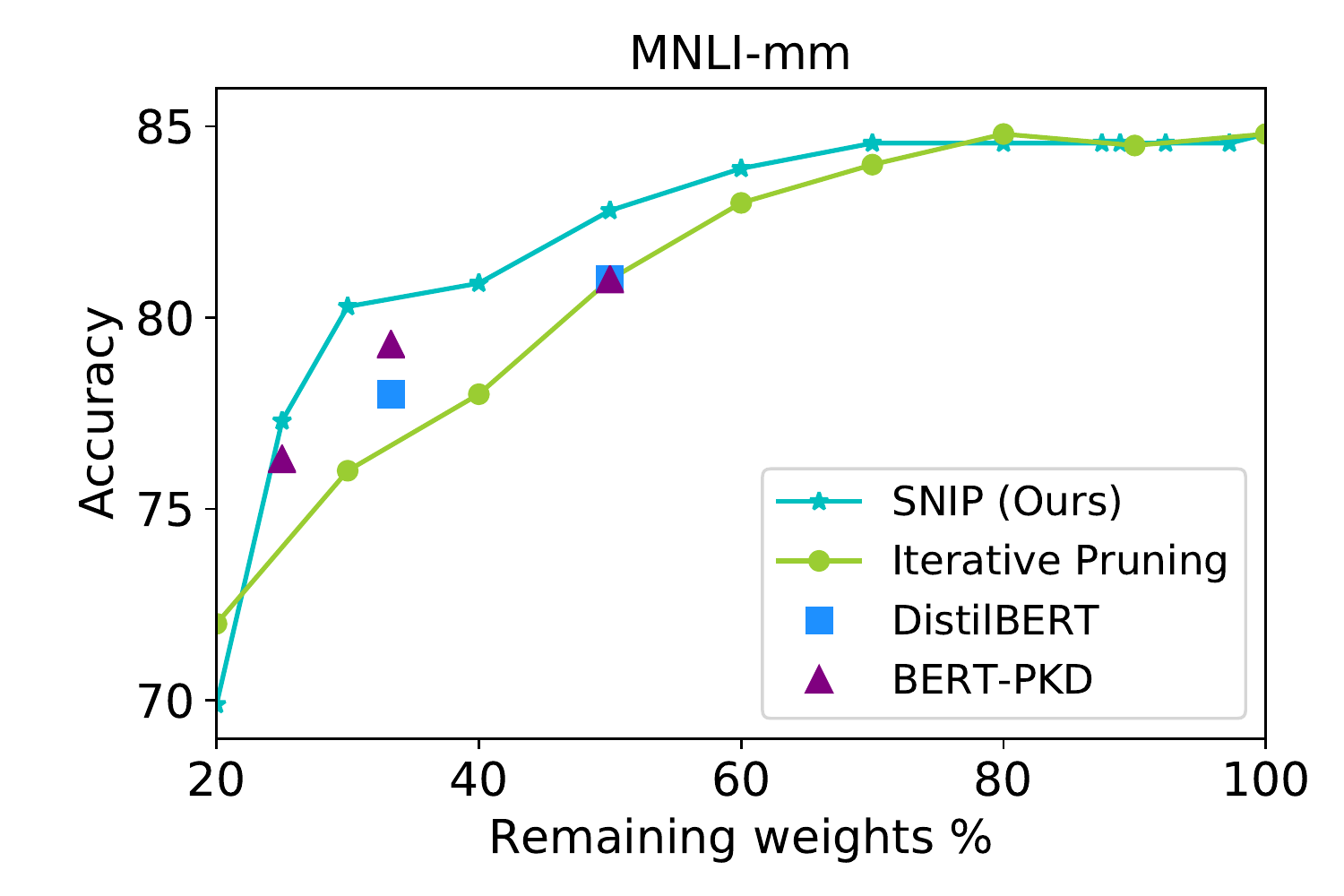}
    \caption{Comparison of selected knowledge distillation methods (DistilBERT \cite{sanh2019distilbert} and BERT-PKD \cite{sun2019patient}) and iterative pruning methods (Iterative Pruning \cite{guo2019reweighted} and our proposed method) in terms of accuracy at various compression rate using MNLI test set. knowledge distillation methods require re-distillation from the teacher to get each single data point, whereas iterative pruning methods can produce continuous curves at once.}
    \label{fig:example}
\end{figure}

Knowledge distillation methods require the specification of a student network with a smaller architecture, which often has to be identified by a tedious sequence process of trial-and-error based decisions.
By comparison, the iterative pruning methods gradually prune the redundant model weights or layers from the full-size model, and provide a full picture of the trade-off between the task performance and the model size with a single training process, as illustrated in Figure \ref{fig:example}. This allows the iterative pruning methods to easily determine the most compact architecture given a required level of model performance.

However, many of the existing pruning methods rely on classic regularizers that act on the individual weights by penalizing them to zero \cite{guo2019reweighted, sanh2020movement}. As a result, the pruned model tends to maintain the same architecture as the original model despite the reduced parameter count, which does not practically lead to an improvement in inference latency \cite{wen2016learning}.

This leads to a question: is it possible to perform more structured pruning on a Transformer model to modify its model architecture (e.g., reducing width and depth)?

To this end, we notice that many previous works have suggested that the learned Transformer models often have much redundancy \cite{tenney2019bert, liu2019linguistic, jawahar2019does, kovaleva2019revealing}. For example, \citet{michel2019sixteen, voita2019analyzing} found that most of the attention heads in a Transformer model can be removed without significantly impacting accuracy, and \citet{tenney2019bert} found that while the earlier layers and the later layers in a BERT model play clear roles in extracting either low-level or task-specific linguistic knowledge, the roles of the intermediate layers are less important. These observations have motivated the idea that Transformer models may exhibit considerable \textit{structural redundancy} - i.e., some layers can be removed during the training without harming the final performance.

In this paper, we propose a \emph{structured} pruning method that reduces the architecture of a Transformer by directly targeting its sub-modules as a whole - for example, a single attention head, an attention module, a feed-forward subnetwork, etc.
We take an approach we call \textit{spectral-normalized identity prior} (SNIP), which imposes a function-level prior centered around the identity function on the aforementioned modules.

Specifically, we take the advantage of the residual blocks ($\mathcal{F}(x) + x$) within a Transformer layer and compress them to \emph{strict identity mappings} \cite{yu2018learning} by identifying the residual blocks whose nonlinear mapping's absolute values ($|\mathcal{F}(x)|$) mostly fall below a threshold $\epsilon$. 
With this strategy, the weights of the Transformer model can still be under-regularized when using simple $L_1$ or $L_2$ based regularizers, which cause the distribution of the post-activation values prone to be noisy even after layer normalization \cite{ba2016layer}.
To address this issue, we further leverage \emph{spectral normalization} \cite{miyato2018spectral} to stabilize the distribution of the post-activation values by regularizing the largest singular value of the weight matrices.

We use BERT \cite{devlin2018bert} as a case study in this paper.
Across multuple tasks in the GLUE benchmark \cite{wang2018glue}, SNIP improves the performance over the state-of-the-art by $0.5$ to $1.0\%$ on average at 50\% compression ratio. \footnote{Open-source code can be found at \url{https://github.com/google-research/google-research/tree/master/snip}}
We also show that spectral normalization results in more sparse and regulated layer mappings during pre-training.
We compare the remaining model components across the tasks at a fixed compression ratio in an ablation study, and show that the remaining components are similar but not identical.

Our contributions are three-fold: First, we introduce \textit{identity-inducing prior}, a structured pruning approach that imposes identity-inducing regularization on the Transformer mappings as a whole rather than its individual weights.
Second, we show that through a novel combination with the \textit{spectral normalization} regularization, the resulting \textit{spectral-normalized identity prior} (SNIP) leads to well-regularized weight distribution and sparse layer mappings in a BERT model. 
Finally, we conduct thorough experiments to validate the SNIP approach over 5 standard NLU tasks. Our results suggest that different model components in a Transformer play critically different roles across tasks, suggesting the importance of performing task-specific pruning to obtain an architecture that is the most suitable for the target task.
\section{Related Work}
\paragraph{Pre-trained Language Model Compression}
The major existing efforts to compress pre-trained language models such as BERT include knowledge distillation \cite{ba2014deep, hinton2015distilling} and pruning \citep{iandola2016squeezenet, veit2017convolutional}.

The knowledge distillation approach enables the transfer of knowledge from a large \textit{teacher} model to a smaller \textit{student} model. Such attempts have been made to distill BERT models, e.g., DistilBERT \cite{sanh2019distilbert}, BERT-PKD \cite{sun2019patient}, Distilled BiLSTM \cite{tang2019distilling}, TinyBERT \cite{jiao2019tinybert}, MobileBERT \cite{sun2020mobilebert}, etc.
All of these methods require carefully designing the student architecture. Furthermore, to choose which intermediate results that the student model can learn from, e.g., the outputs of each layer, the attention maps, is still under discussion.

Similar to other pruning-based methods, our method can iteratively remove the least important weights or connections, explore the full spectrum of trade-offs, and find the best affordable architecture in one shot.

Many language representation model pruning methods focus on individual components of the weight matrices. For example, \citet{guo2019reweighted} integrates reweighted $L_1$ minimization with a proximal algorithm to search sparsity patterns in the model; \citet{gordon2020compressing} uses magnitude weight pruning, which compresses the model by removing weights close to 0; \citet{sanh2020movement} applies deterministic first-order weight pruning method where both weights with low and high values can be pruned. A very few works try structured weight pruning, e.g., \citet{wang2019structured} proposes a structured pruning approach based on low-rank factorization and augmented Lagrangian $L_0$ norm regularization. On the other hand, there also exist works that prune a coherent set of sub-modules in the Transformer model. For example, 
\citet{michel2019sixteen} and \citet{voita2019analyzing} propose to prune individual attention heads either manually via head importance score, or automatically via a relaxed $L_0$ regularization. \citet{fan_reducing_2020} applies random pruning to the entire layers. In contrast, our method allows finer-grained structured pruning on Transformer modules (i.e., both attention heads and feed-forward layers) and propose to improve the mathematical property of a Transformer (i.e., Lipschitz condition) for more effective pruning.

Other compression approaches include weight sharing \cite{liu2019roberta}, quantization \cite{zafrir2019q8bert, shen2019q} and neural architecture search \cite{chen2020adabert}, but are not within the discussion of this paper. We refer interested readers to \citet{ganesh2020compressing} for further details.

\paragraph{Applications of Spectral Normalization}

Spectral normalization is first proposed for generative adversarial network (GAN) as a regularization technique to stabilize the discriminator training \citep{miyato2018spectral}. It was later applied to improve the performance of the other types of generative neural networks \citep{zhao_adversarially_2018, behrmann_invertible_2019}, and was analyzed theoretically in the context of adversarial robustness and generalization \citep{farnia_generalizable_2018}.

Spectral normalization regularizes the Lipschitz condition of the model mappings and is known to benefit model generalization under both the classic and the adversarial settings \citep{sokolic2017robust, cisse2017parseval, oberman2018lipschitz, neyshabur2017exploring}. In this paper, we will explore the benefit of spectral regularization for improving the effectiveness of pruning.
\section{Methods}

In this section, we first briefly review the basic Transformer layers in \citet{vaswani2017attention} (\ref{sec:background}). We then introduce our identity prior into Transformer's residual connections using $\epsilon$ threshold (\ref{sec:identity_mapping}). In section \ref{sec:sepctral-normalization}, we give mathematical foundations to the spectral normalization and show how it could help with our identity prior. Finally, to put it all together, we establish our structured iterative pruning methods for BERT fine-tuning (\ref{sec:iterative-pruning}).

\subsection{Background: Transformer Layer}
\label{sec:background}
Transformer-based models are usually comprised of a stack of Transformer layers. A Transformer layer takes on a sequence of vectors as input, first passes it through a (multi-head) self-attention sub-layer, followed by a position-wise feed-forward network sub-layer.

\paragraph{Self-attention sub-layer}
The attention mechanism can be formulated as querying a dictionary with key-value pairs. Formally, 
\begin{align*}
    \textrm{Att}(Q, K, V)=\textrm{softmax}(QK^T/\sqrt{d_H}) \cdot V
\end{align*}
where $d_H$ is the dimension of the hidden representations. $Q$, $K$, and $V$ represent query, key, and value. The multi-head variant of attention (MHA) allows the model to jointly attend to information from different representation sub-spaces, defined as
\begin{align*}
    \textrm{MHA}(Q, K, V) &= [\textrm{head}_1, \ldots, \textrm{head}_A] W^O \\
     \textrm{head}_k &= \textrm{Att}(QW_k^Q, KW_k^K, VW_k^V)  
\end{align*}

\noindent where $[\cdot,\cdot]$ is the concatenation operator, $W_k^Q\in \mathbb{R}^{d_H \times d_K}$, $W_k^K\in \mathbb{R}^{d_H\times d_K}$, $W_k^V\in \mathbb{R}^{d_H\times d_V},$ and $W^O\in \mathbb{R}^{Hd_V \times d_H}$ are projection parameter matrices, $A$ is the number of heads, and $d_K$ and $d_V$ are the dimensions of key and value. 

\paragraph{Position-wise FFN sub-layer}
In addition to the self-attention sub-layer, each Transformer layer also contains a fully connected feed-forward network, which is applied to each position separately and identically. This feed-forward network consists of two linear transformations with an activation function $\sigma$ in between. Specially, given vectors $\mathbf{x}_1, \ldots, \mathbf{x}_n$, a position-wise FFN sub-layer transforms each $\mathbf{x}_i$ as $\text{FFN}(\mathbf{x}_i) = \sigma(\mathbf{x}_iW_1+b_1) W_2 + b_2$, where $W_1, W_2, b_1$ and $b_2$ are parameters.

We should also emphasize that a residual connection \cite{he2016identity} and a layer normalization \cite{ba2016layer} are applied to the output of both MHA or FNN sub-layers. The residual connection plays a key role in learning strict identity mapping (detailed in Section \ref{sec:identity_mapping}), while layer normalization and spectral normalization (detailed in Section \ref{sec:sepctral-normalization}) together ensure regulated magnitude of activation outputs for improved pruning stability.

\subsection{Identity-inducing Prior for Transformer}
\label{sec:identity_mapping}

The design of residual connection can provide us with a promising way to find identity mappings for the Transformer model. Specifically, residual connection \cite{he2016identity} can be formalized as $\mathcal{H}(\mathbf{x}) = \mathcal{F}(\mathbf{x}) + \mathbf{x}$, where $\mathcal{F}$ could be either $\textrm{MHA}$ or $\textrm{FFN}$ and $\mathcal{H}$ is the sub-layer output.
As illustrated in \citet{he2016deep}, if an identity mapping is optimal, it is easier to push the residual to zero than to fit an identity mapping by a stack of traditional non-linear layers.

We leverage $\epsilon$-ResNet \cite{yu2018learning}, a strict identity mapping mechanism that sparsifies the layer output by inducing a specific threshold $\epsilon$ as the \textit{identity prior}.
Specifically, we turn the residual connection $\mathcal{H}(\mathbf{x})$ into $\mathcal{S}_\epsilon(\mathcal{F}(\mathbf{x})) + \mathbf{x}$, where 
\begin{equation*}
\mathcal{S}_\epsilon(\mathbf{v}) = 
\left\{
\begin{array}{ll}
    0 &  \textrm{if } |\mathbf{v}_i|<\epsilon, \forall i \in {1, ..., |\mathbf{v}|},\\
    \mathbf{v} & \textrm{otherwise,}
\end{array}
\right.
\end{equation*}

\noindent where $\mathbf{v}_i$ is the $i$-th element of the vector $\mathbf{v}$.
Here a sparsity-promoting function $\mathcal{S}$ is applied to dynamically discard the non-linearity term based on the activations. When all the responses in the non-linear mapping $\mathcal{F}(\mathbf{x})$ is below a threshold $\epsilon$, then $\mathcal{S}(\mathcal{F}(\mathbf{x}))=0$, otherwise, the original mapping $\mathcal{S}(\mathcal{F}(\mathbf{x}))=\mathcal{F}(\mathbf{x})$ was used as the standard residual network.

To implement $\mathcal{S}$, we put an extra binary gate layer $t_\epsilon$ upon $\mathcal{F}(\mathbf{x})$ by stacking additional rectified linear units (ReLU), following \citet{srivastava2015highway}. In particular,
\begin{equation*}
    t_\epsilon (\mathbf{v}) = 1 - \textrm{ReLU}(1-L \max\limits_{1 \le i \le |\mathbf{v}|}\textrm{ReLU}(|\mathbf{v}_i| - \epsilon))
\end{equation*}

\noindent where $L$ refers to a very large positive constant (e.g., $1e5$ in our experiments). Then, $\mathcal{S}_\epsilon(\mathcal{F}(\mathbf{v}))$ has the following form:
\begin{equation*}
    \mathcal{S}_\epsilon(\mathcal{F}(\mathbf{x})) = t_\epsilon (\mathcal{F}(\mathbf{x})) \mathcal{F}(\mathbf{x})
\end{equation*}

Recall that each layer of Transformer consists of two residual blocks, namely, the self-attention sub-layer and the position-wise FFN sub-layer. We apply the $\epsilon$ network directly to the residual block in the FFN sub-layer, i.e., 
\begin{align*}
    \mathcal{H}_\textrm{FFN}(\mathbf{x}) = \mathcal{S}_{\epsilon_\textrm{FFN}}(\textrm{FFN}(\mathbf{x})) + \mathbf{x},
\end{align*}

\noindent When applying it to the attention sub-layer, we place $\mathcal{S}$ to each single attention head, which allows us to prune any subset of attention heads, i.e., 
\begin{align*}
    \mathcal{H}_\textrm{ATT}(\mathbf{x}) = \sum_{i=1}^A \mathcal{S}_{\epsilon_\textrm{ATT}}(\textrm{head}_i W^O_i) + \mathbf{x}
\end{align*}

\noindent Here, $W^O_i$ is the output weight assign to the $i$-th attention head, i.e., $W^O=[W^O_I, W^O_2, ..., W^O_A]$. If $\mathcal{S}_{\epsilon_\textrm{ATT}}(\textrm{head}_i W^O_i) = 0$, this means that the $i$-th attention head does not contribute to the output of the attention layer and thus could be pruned out.

\begin{figure*}[tb]
    \begin{minipage}[t]{0.5\textwidth}
    \centering
        \includegraphics[width=1.0\textwidth]{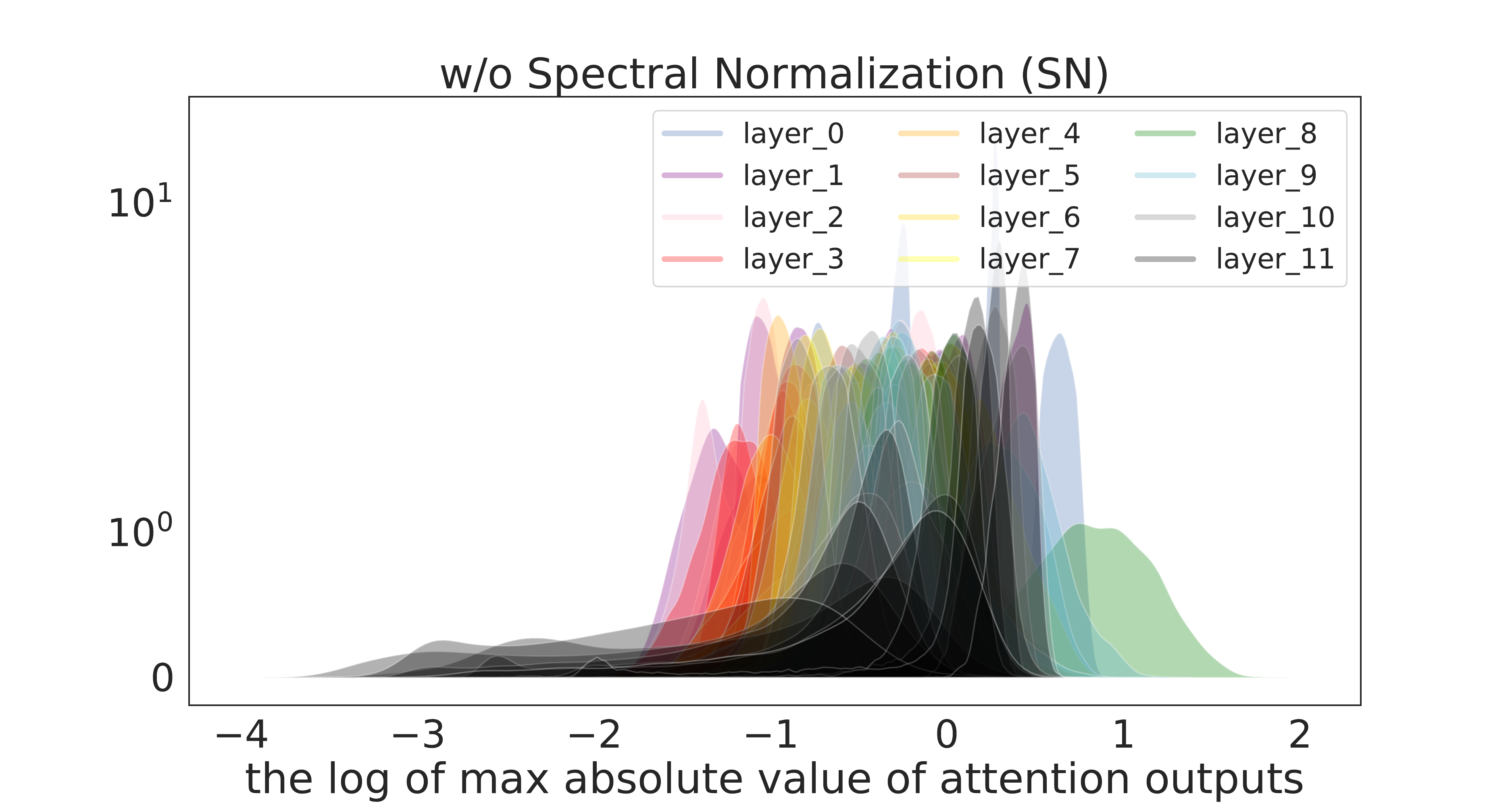}
    \end{minipage}%
    \begin{minipage}[t]{0.5\textwidth}
    \centering
        \includegraphics[width=1.0\textwidth]{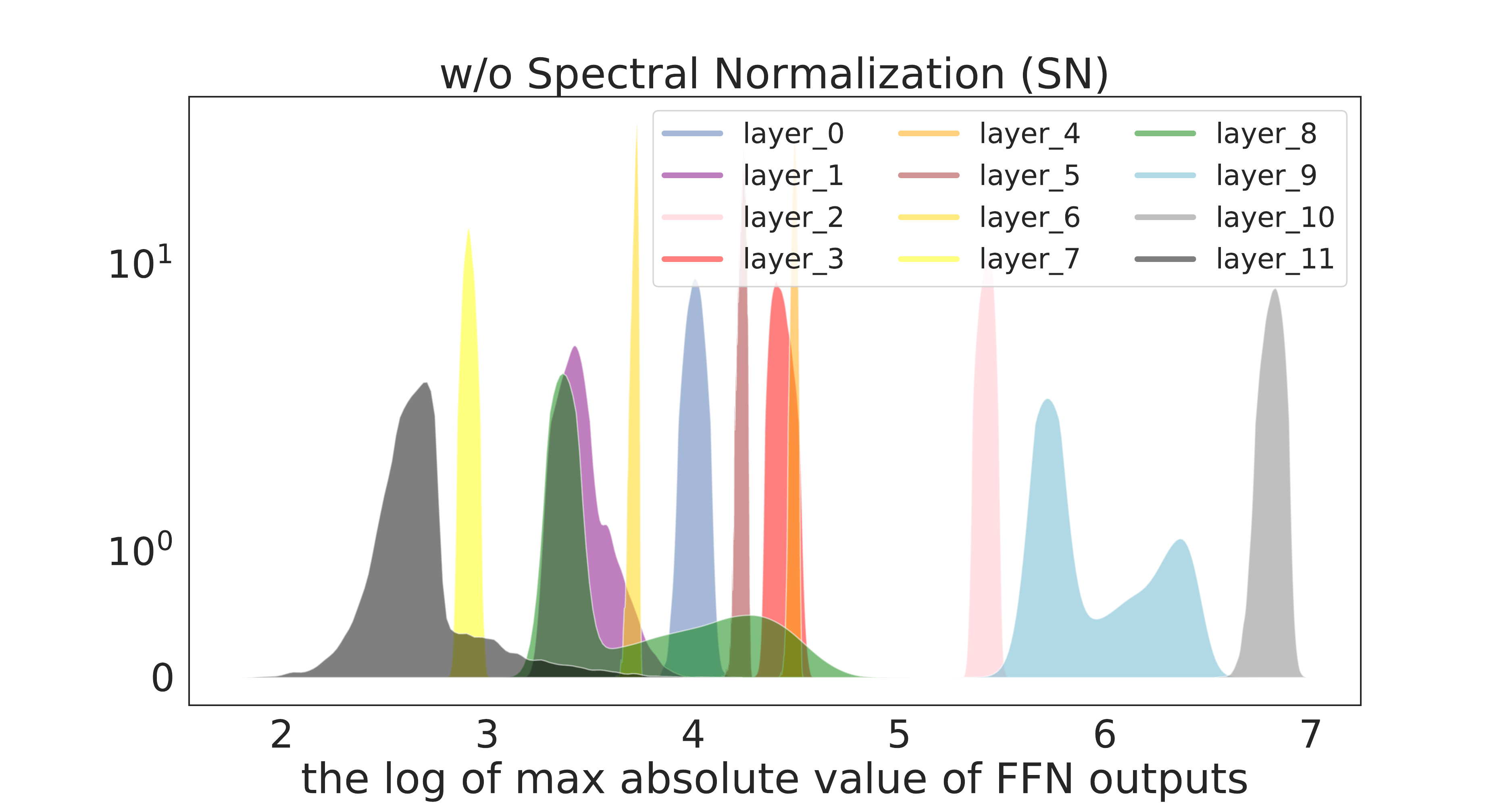}
    \end{minipage}
    \\
    \begin{minipage}[t]{0.5\textwidth}
    \centering
        \includegraphics[width=1.0\textwidth]{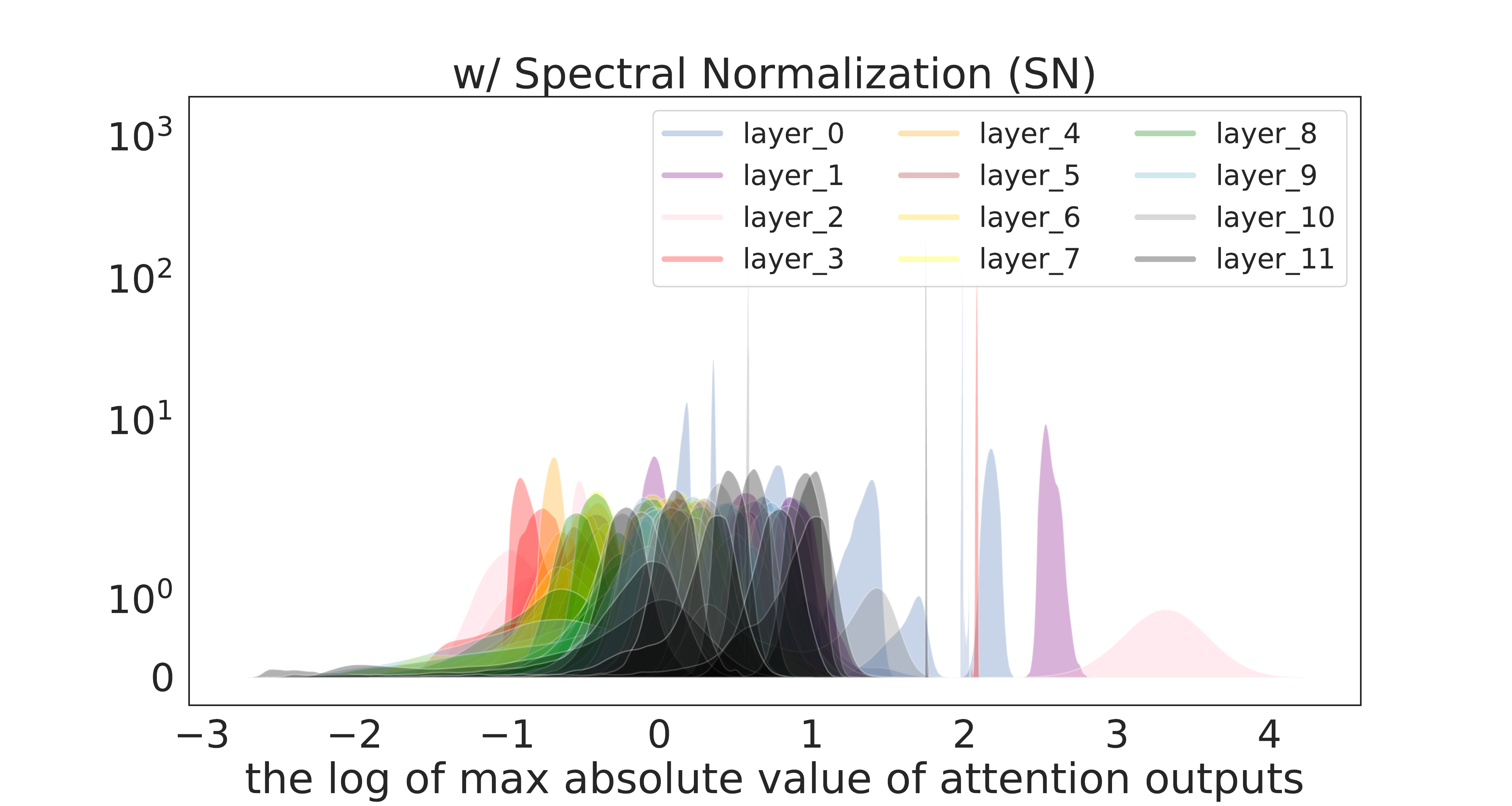}
    \end{minipage}%
    \begin{minipage}[t]{0.5\textwidth}
    \centering
        \includegraphics[width=1.0\textwidth]{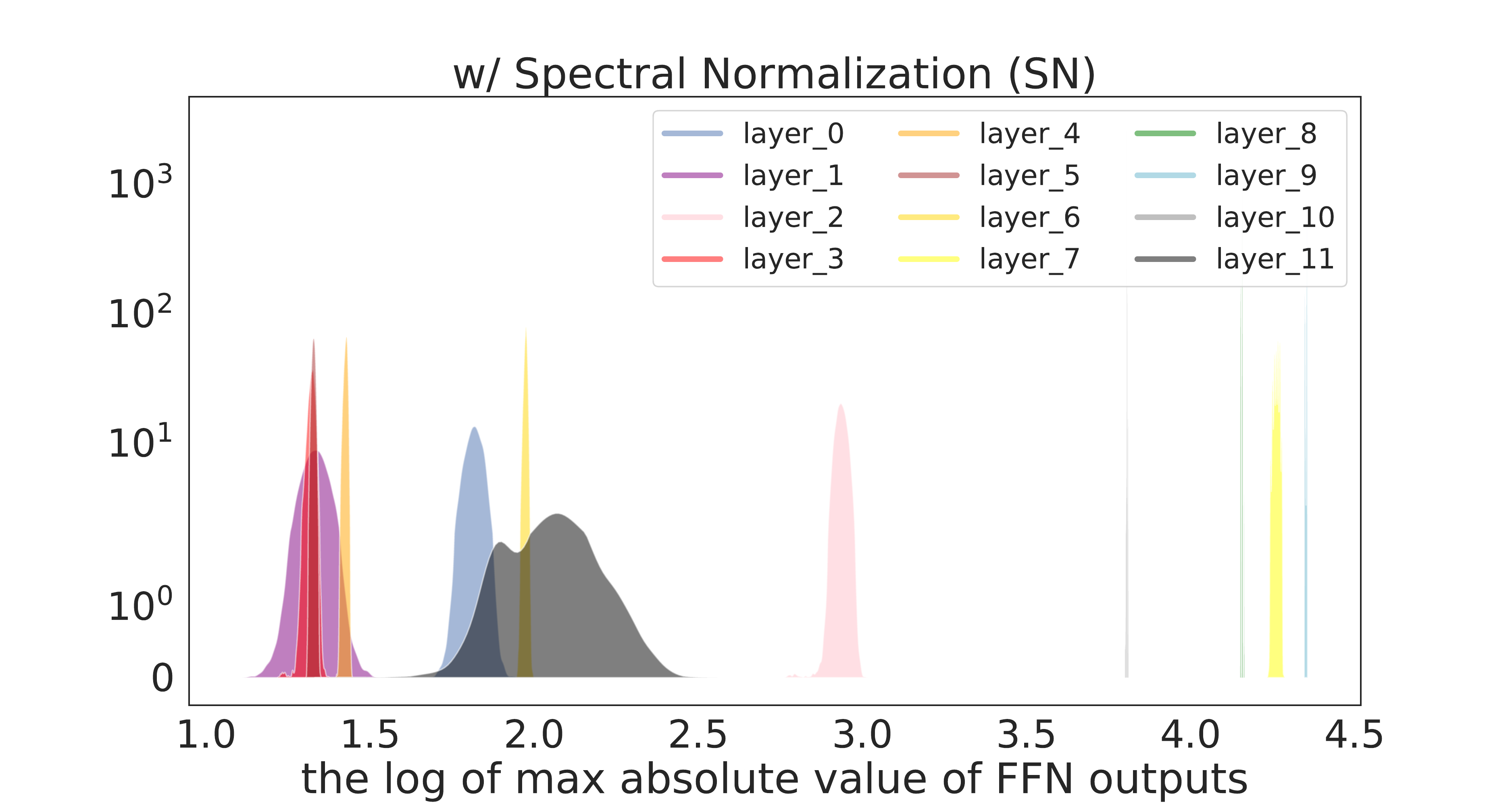}
    \end{minipage}
    \caption{Distribution of the max absolute value of the attention/FFN outputs on the Stanford Sentiment Treebank (SST) training set, based on model BERT$_\textrm{BASE}$ (w/o SN) and spectral normalized BERT$_\textrm{BASE}$ (w/ SN). For attention layers, different heads are plotted separately but with the same color in the same layer for clarity. After SN, the norm distribution $P(|\mathcal{F}_k(\mathbf{x})|)$ shows a better separation (bottom v.s. top), and the norms distribution for the FFN layers are stabilized within a much smaller range (bottom right v.s. top right).}
    \label{fig:output-distribution}
\end{figure*}

In our experiment, we set different values to $\epsilon_\textrm{ATT}$ and $\epsilon_\textrm{FFN}$, since the absolute outputs of attention and FFN layers lay in different scalars, as illustrated in Figure \ref{fig:output-distribution}.

\subsection{Spectral Normalization}
\label{sec:sepctral-normalization}

The sparsity-inducing function $\mathcal{S}_\epsilon(\mathcal{F}(\mathbf{x}))$ in the $\epsilon$-ResNet has been found to work well for randomly initialized neural network, where the initial weight matrix of the non-linear mappings for all layers was distributed within a consistent range, and thus facilitates a natural separation between the function norms $|\mathcal{F}|$ of the important and unimportant non-linear mappings in the residual blocks during training \citep{yu2018learning}. This is, however, not the case for the weight distribution of a pre-trainined model like BERT, where the weight distributions between different layers have already diverged during pre-training, which is likely due to the specialization of layer functionalities under the masked language modeling (MLM) training \citep{tenney2019bert}.

\begin{table}[tb]
    \centering\scalebox{0.8}{
    \begin{tabular}{lcccccc}
    \toprule
        \textbf{SN}                  & \textbf{MLM} & \textbf{SST-2} & \textbf{QQP} & \textbf{MRPC} & \textbf{QNLI} & \textbf{MNLI}\\\midrule
        w/o                & 71.93       & 92.7        & 90.6        & 90.9         & 91.6         & 84.7           \\
        w/ & 74.16       & 91.9        & 90.1        & 90.8         & 91.6         & 85.1           \\
    \bottomrule
    \end{tabular}}
    \caption{Evaluate the performance of BERT$_\textrm{BASE}$ (w/o SN) and the spectral normalized BERT$_\textrm{BASE}$ (w/ SN, using $\lambda(W)=5$), respectively masked language modeling accuracy on pre-training data and accuracy of fine-tuning on 5 natural language understanding tasks (details could be found in Section \ref{sec:datasets}).
    }
    \label{tab:performance-on-different-norm}
\end{table}

Indeed, in our preliminary experiments, we observed that the proposed identity-inducing prior $\epsilon$ is not effective for a BERT model initialized from a classic pre-training checkpoint. As shown in Table \ref{tab:performance-on-different-norm}, we found the distribution of the function norms $P(|F(\mathbf{x})_i|)$ for the attention layers to be densely clustered within a small range ($(0, 2)$) and with no clear separation between the function norm for the important and unimportant non-linear residual mappings. On the other hand, the norm distributions for different FFN layers were found to vary wildly, creating challenges for selecting a proper set of $\epsilon$'s in practice.

The above observations motivate us to identify an effective method to stabilize the norm distributions of BERT model layers. In this work, we consider \textit{spectral normalization} (SN), an approach that directly controls the Lipschitz norm of a non-linear mapping $\mathcal{F}$ by regularizing the spectral behavior of its weight matrices \citep{miyato2018spectral}. 

Specifically, for a weight matrix $W$, its spectral norm $\lambda(W)$ is defined as its largest singular value:
\begin{align*}
    \lambda(W) = 
    \underset{x\neq0}{\textrm{max}}\frac{||W\mathbf{x}||_2}{||\mathbf{x}||_2}.
\end{align*}

\noindent We say a function $\mathcal{F}$ is $L$-Lipschitz if $|\mathcal{F}(\mathbf{x}_1)-\mathcal{F}(\mathbf{x}_2)|/||\mathbf{x}_1 - \mathbf{x}_2|| \leq L$, for all possible $(\mathbf{x}_1, \mathbf{x}_2)$ pairs from the feature space, and we call the smallest possible $L$ the Lipschitz norm of $\mathcal{F}$, denoted as $|\mathcal{F}|_{Lip}$. 
Consequently, for a neural network mapping $\mathcal{F}(\mathbf{x}) = \sigma(W\mathbf{x} + b)$ with an contractive activation function $\sigma$,
its Lipschitz norm is upper-bounded by $\lambda(W)$ \citep{miyato2018spectral}:
\begin{equation*}
    |\mathcal{F}|_{Lip} = |\sigma(W \mathbf{x} + b)|_{Lip} \leq |W \mathbf{x} + b|_{Lip} \leq \lambda(W).
\end{equation*}

For BERT models, since the layer input $\mathbf{x}$ follows a  distribution of zero mean and unit variance due to layer normalization \cite{ba2016layer}, a nonlinear mapping's $L_1$ norm $|\mathcal{F}|$ is roughly proportional to its Lipschitz norm $|\mathcal{F}|_{Lip}$, which is controlled by $\lambda(W)$.
Therefore, we are able to have a better control of the maximum of $|\mathcal{F}(\mathbf{x})|$ for identifying a good $\epsilon$. Furthermore, the regularization is achieved by that the layer weights are simply divided by their corresponding spectral norm in SN, i.e., $\hat{W} = W / \hat{\lambda}(W)$, adding no additional trainable parameter to the original model.

\begin{figure*}[tb]
    \begin{minipage}[t]{0.33\textwidth}
        \includegraphics[width=5.5cm]{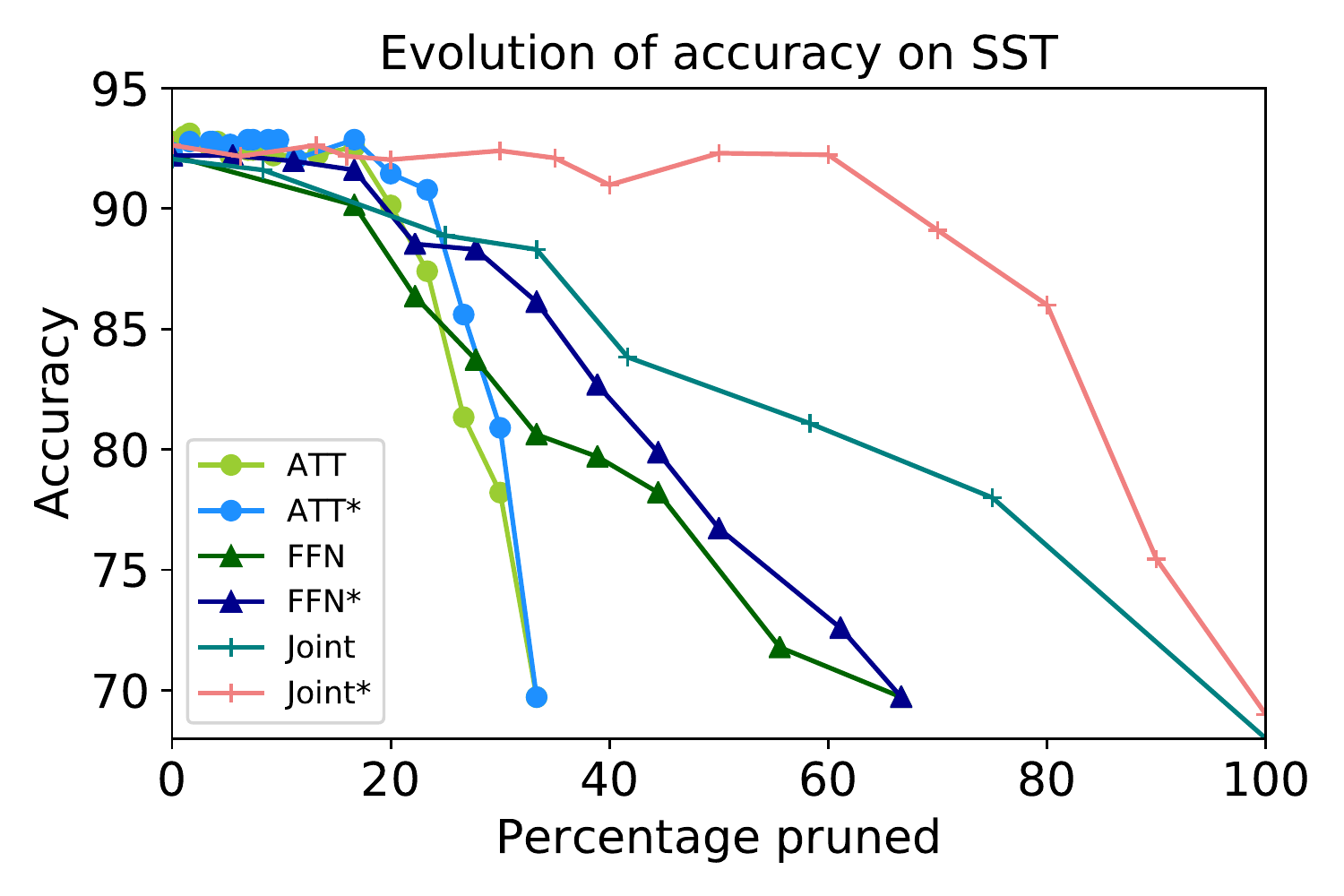}
    \end{minipage}%
    \begin{minipage}[t]{0.33\textwidth}
        \includegraphics[width=5.5cm]{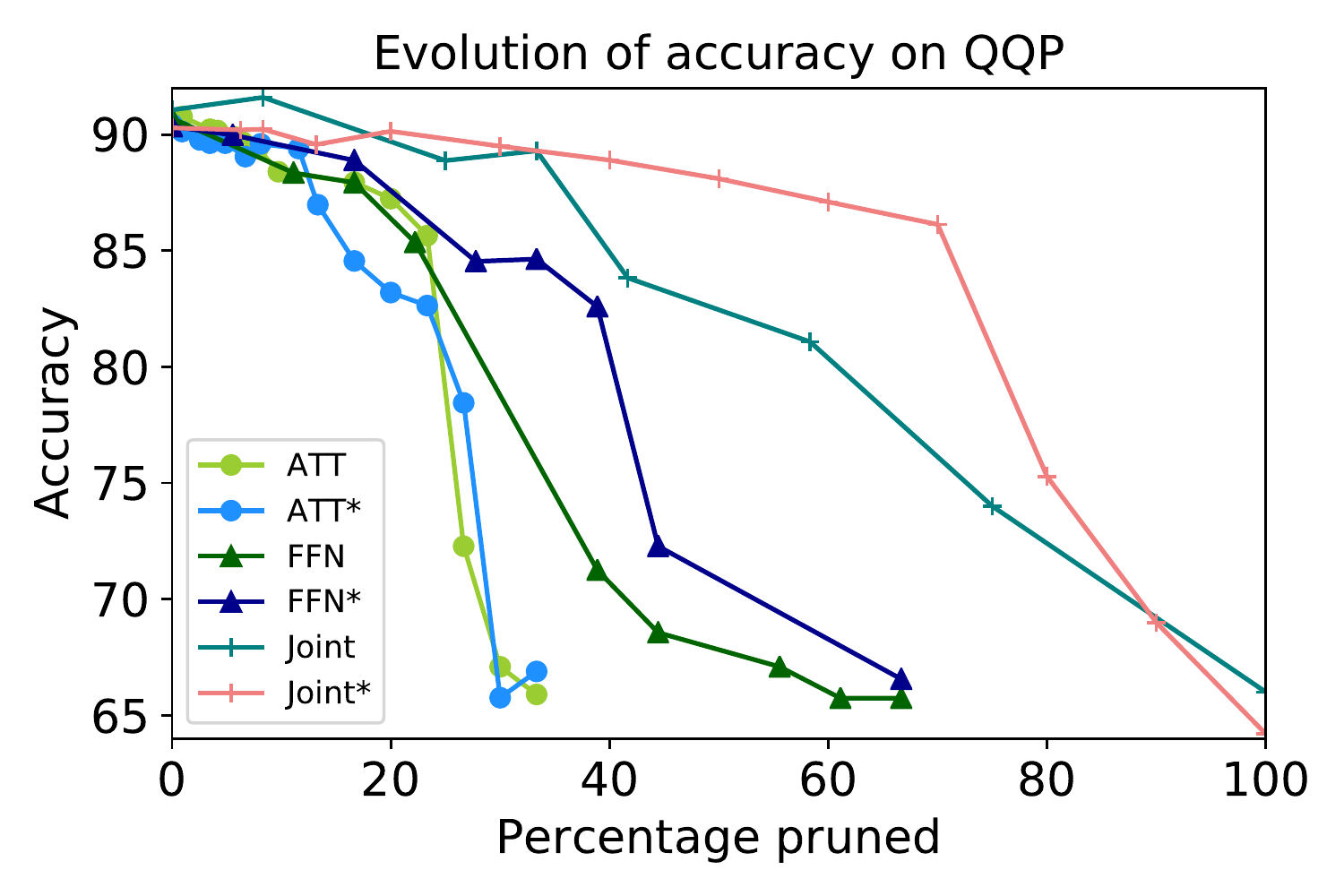}
    \end{minipage}
    \begin{minipage}[t]{0.33\textwidth}
        \includegraphics[width=5.5cm]{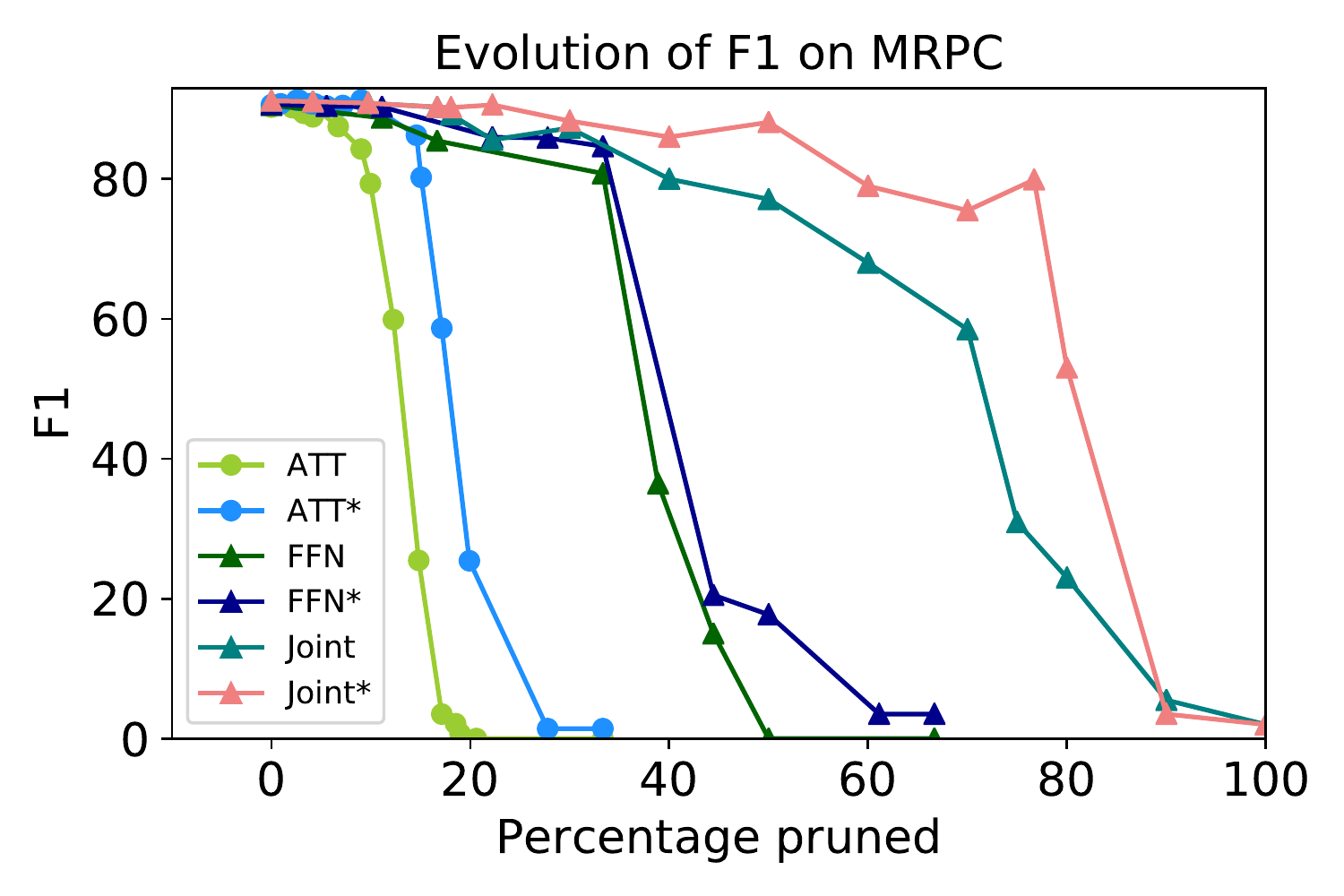}
    \end{minipage}%
    \\
    \begin{minipage}[t]{0.5\textwidth}
        \includegraphics[width=5.5cm, right]{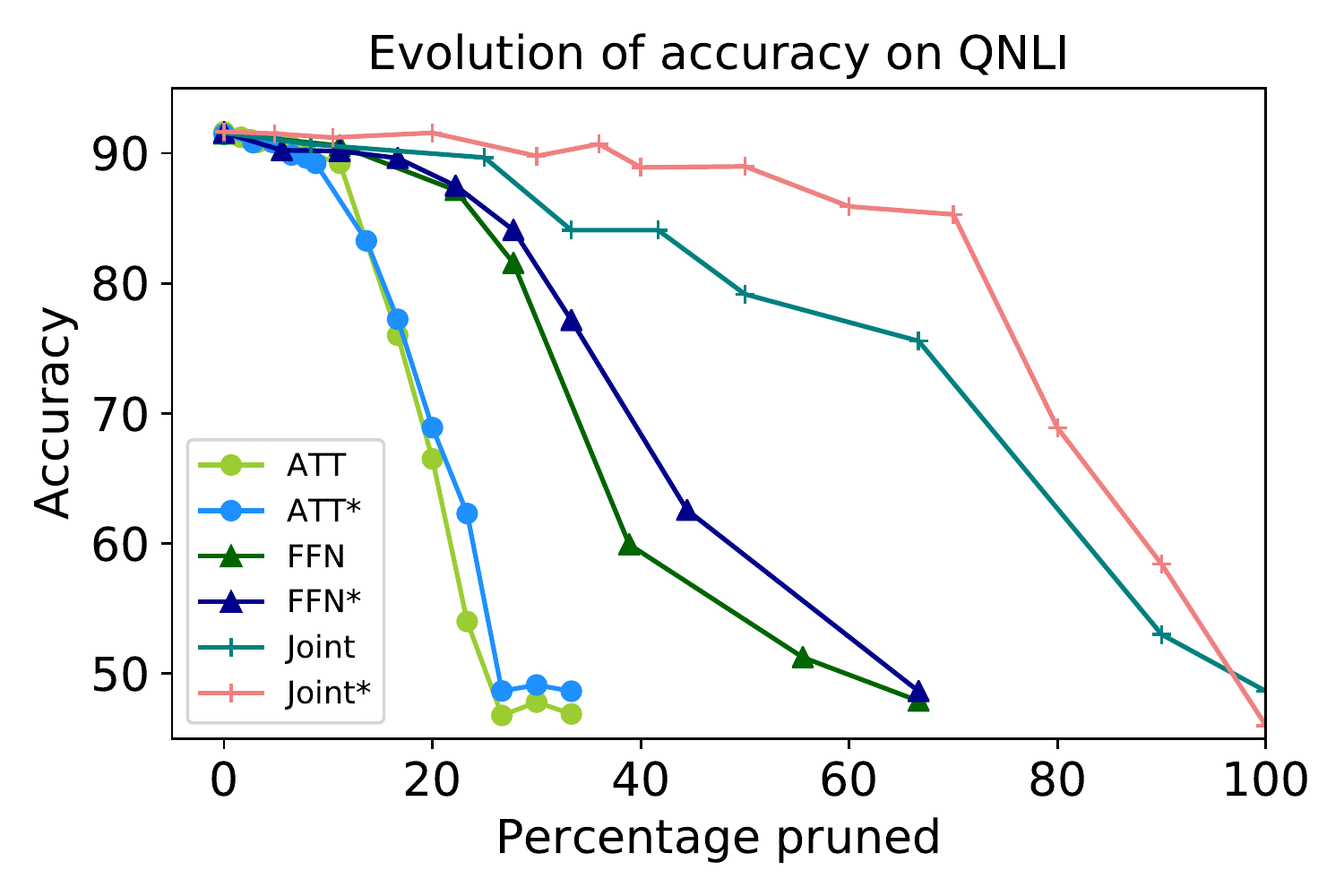}
    \end{minipage}%
    \begin{minipage}[t]{0.5\textwidth}
        \includegraphics[width=5.5cm, left]{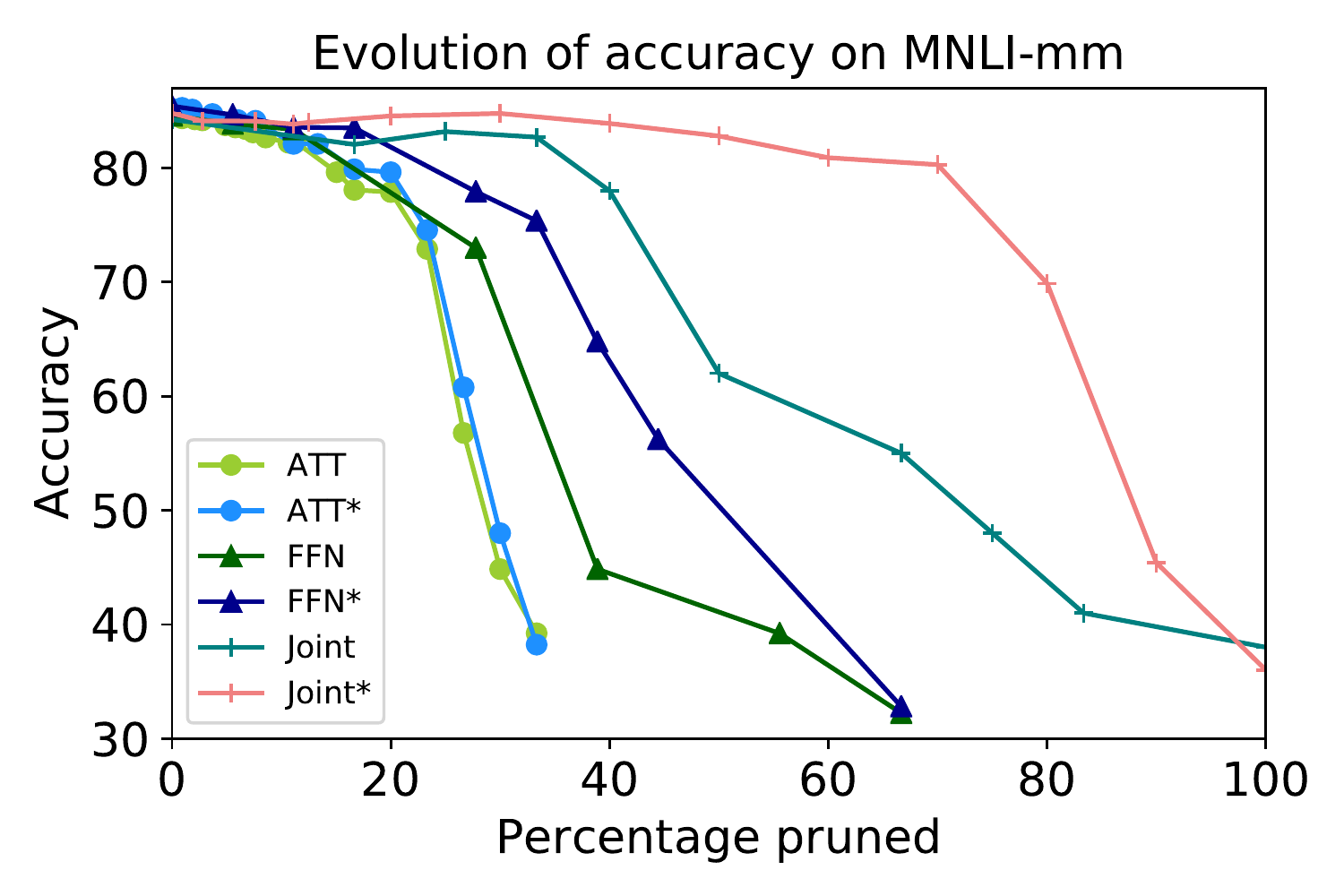}
    \end{minipage}
\caption{Evolution of performance when separately pruning attention heads (ATT) and FFN layers (FFN), and joint pruning ATT and FFN (Joint) on 5 GLUE dev sets. * means applying spectral normalization. The percentage pruned (x-axis) is calculated based on the parameters of attention and FFN sub-layers as the full set (excluding embedding, pooler and classifier layers). Joint pruning with spectral normalization (Joint*) could prune more parameters than the original BERT (Joint) to preserve the same accuracy.}
\label{fig:results}
\end{figure*}

We apply SN to both pre-training and fine-tuning of the BERT model, and on the weights in both the attention and the FFN layers. As shown in Table \ref{tab:performance-on-different-norm}, compared to the original BERT$_\textrm{BASE}$ without SN, adding SN to a BERT$_\textrm{BASE}$ model has resulted in improved pre-training performance and competitive fine-tuning performance. To illustrate the effect of SN on the distribution of function norms of the BERT model, we plot $|\mathcal{F}(\mathbf{x})|$ for the training set of the Stanford Sentiment Treebank in Figure \ref{fig:output-distribution}. As shown, after SN, the norm distribution $P(|\mathcal{F}_k(\mathbf{x})|)$ between the important and the unimportant nonlinear mappings shows a better separation (Figure \ref{fig:output-distribution}, bottom), the norms distribution for the FFN layers are now stabilized within a much smaller range (Figure \ref{fig:output-distribution}, bottom right).

\subsection{Structured Iterative Pruning}
\label{sec:iterative-pruning}
We use a simple pruning method that greedily and iteratively prunes away attention heads and FFN layers to avoid impractical combinatorial search, where two dynamic estimations are conducted for $\epsilon$ and model architecture respectively.
One iteration contains four substeps: 
\begin{enumerate}
    \item Estimate $\epsilon$ given current model architecture and training data. Specifically, 
    We sort the attention heads and FFN layers by their mean activation outputs, and set $\epsilon$ to the $k$-th smallest mean activation.
    Larger $k$ leads to more mappings being pruned in one iteration, which makes the retraining more difficult to recover the performance, but leads to fewer pruning iterations.
    
    \item Train the model with identity-inducing prior by using the selected $\epsilon$.
    
    \item Estimate a smaller architecture given current $\epsilon$ and training data.
    Specifically, we estimate the module usage by counting the number of times each residual block has been learned to become a strict identity mapping across mini-batches in the training set.
    We prune residual blocks whose usage rate below a threshold $\theta$.
    When a residual block produces a negligible response, the $\epsilon$ function will start producing $0$ outputs. As a result, the weights in this block will stop contributing to the cross-entropy term. Consequently, the gradients will only be based on the regularization term and lead to weight collapse.
    
    \item Retrain the model with the pruned residual blocks completely removed from the architecture.
    This is critical --- if the pruned network is used without retraining, accuracy is significantly impacted. 
    Also, during retraining, it is better to retain the weights from the initial training phase for the connections that survived pruning than it is to re-initialize them \cite{han2015learning}.
\end{enumerate}
\begin{table*}[tb]
    \centering
    \scalebox{0.76}{
    \begin{tabular}{lrccccc}
    \toprule
        \textbf{Model}                            & \textbf{FLOPS} & \textbf{ SST-2$_{acc}$} & \textbf{QQP$_{acc}$} & \textbf{MRPC$_{F_1}$} & \textbf{QNLI$_{acc}$} & \textbf{MNLI-mm$_{acc}$}\\\hline
        BERT$_\textrm{BASE}$                      & 22.5B          & \ \ -0.0\%/92.7        & \ \ -0.0\%/90.6       & \ \ -0.0\%/90.9        & \ \ -0.0\%/91.6        & \ \ -0.0\%/84.7 \\
        TinyBERT$_6$ w/o DA \cite{jiao2019tinybert}\protect\footnotemark & 1.2B         & -                      & -                     & -50.0\%/86.0           & -                      & \textbf{-50.0\%/84.4}\\
        DistilBERT$_6$ \cite{sanh2019distilbert} & 11.3B          & -50.0\%/91.3           & -50.0\%/88.5          & -50.0\%/87.5           & -50.0\%/89.2           & -50.0\%/82.2 \\
        BERT-PKD$_6$ \cite{sun2019patient}        & 11.3B          & -50.0\%/91.5           & \textbf{-50.0\%/88.9}          & -50.0\%/86.2           & -50.0\%/89.0           & -50.0\%/81.0 \\
        BERT-PKD$_3$ \cite{sun2019patient}        & 7.6B           & -75.0\%/87.5           & -75.0\%/87.8          & -75.0\%/80.7           & -75.0\%/84.7           & -75.0\%/76.3 \\
        \hline
        FLOP \cite{wang2019structured}           & 15B             & -35.0\%/92.1           & -                     & -35.0\%/88.6           & -35.0\%/89.0           & -           \\
        MvP \cite{sanh2020movement} $\dagger$    & N/A             & -                      & -97.0\%/89.2          & -                      & -                      & -97.0\%/79.7\\\hline\hline
        SNIP (w/ SN)          & 13.2 - 14.5B    & \textbf{-50.0\%/91.8}           & \textbf{-50.0\%/88.9}         & \textbf{-50.0\%/88.1}           & \textbf{-50.0\%/89.5}           & -50.0\%/82.8\\
        SNIP (w/ SN)          & 8.3 - 9.1B      & -75.0\%/88.4           & -75.0\%/87.8          & -75.0\%/81.2           & -75.0\%/84.6           & -75.0\%/78.3\\\hline
        SNIP (w/o SN)                          & 16.8 - 18.2B    & -30.0\%/91.3           & -38.7\%/89.5          & -39.7\%/89.9           & -26.1\%/90.8           & -32.3\%/83.5\\
        SNIP (w/ SN)          & 13.4 - 16.1B    & -56.7\%/91.7           & -40.7\%/89.7          & -46.7\%/89.9           & -36.0\%/90.7           & -39.3\%/83.9\\
    \bottomrule
    \end{tabular}}
    \caption{The compression results including model efficiency (percentage of reduced parameters) and performance from the GLUE dev results, and the MNLI result is evaluate for mismatched-accuracy (MNLI-mm). BERT$_\textrm{BASE}$ indicates the results of the fine-tuned BERT$_\textrm{BASE}$ in our implementation. The number of model parameters includes the attention and FFN sub-layers but excludes the embedding, pooler and classifier layers. The bold numbers indicate the best performance for keeping 50\% of the parameters respectively. Rows 1--5 are knowledge distillation methods, and Rows 6 and 7 are pruning methods, where $\dagger$ means unstructured pruning \protect\footnotemark.}
    \label{tab:best-architecture}
\end{table*}

\section{Experiments}

\subsection{Experimental Settings}
In the experiments, we apply the same architecture and the base settings from the original BERT$_\textrm{BASE}$ \cite{devlin2018bert}, and fine-tune each task independently. More details could be found in Appendix.

\label{sec:datasets}
For pre-training, we use the same data as BERT, which consists of 3.3 Billion tokens from Wikipedia and BooksCorpus \citep{zhu2015aligning}. Similar to the standard BERT practice, we conduct the pre-training only once and from scratch (i.e., no second pre-training). We use dynamic token masking with the masked positions decided on-the-fly instead of during preprocessing. Also, we did not use the next sentence prediction objective proposed in the original BERT paper, as recent work has suggested it dost not improve the scores \cite{yang2019xlnet, liu2019roberta}.

For fine-tuning tasks, we focus on the General Language Understanding Evaluation (GLUE) benchmark \citep{wang2018glue} in the main text since it is a thoroughly studied setting in many pruning/distillation work, thereby allowing comprehensive comparison.
We conduct experiments on the subset of GLUE , classified into three categories:
\begin{enumerate}
    \item Sentiment analysis: Stanford Sentiment Treebank (SST) \cite{socher2013recursive};
    \item Paraphrasing: Quora Question Pairs (QQP) \cite{chen2018quora} and Microsoft Research Paraphrase Corpus (MRPC) \cite{dolan2005automatically};
    \item Natural language inference: Question Natural Language Inference (QNLI) \cite{chen2018quora}, and Multi-genre Natural Language Inference (MNLI) \cite{williams2017broad}. 
\end{enumerate}
The detailed description of downstream tasks could be found in Appendix. The reason for choosing this subset is that we found the variance in performance for those tasks lower than the other GLUE tasks. 

We used the hyperparameters from \citet{clark2020electra} for the most part. Since we run the training iteratively, we set train epoch as 1 for most of the task but 3 for task MRPC consider that the size of the datasets is much smaller than other tasks.

For $\epsilon$ and architecture estimations in Section \ref{sec:iterative-pruning}, we set $k=1$ and $\theta = 0.95$ in our experiment.

\begin{figure*}[tb]
    \centering
    \renewcommand{\arraystretch}{0}
    \begin{tabular}{p{0.005\textwidth}p{0.28\textwidth}p{0.28\textwidth}p{0.28\textwidth}}
            & \multicolumn{1}{c}{\ \ \ \ \ \ SST}   & \multicolumn{1}{c}{\ \ \ \ \ \ MRPC} &\multicolumn{1}{c}{\ \ \ \ \ \ MNLI} \\
     \rotatebox{90}{\ \ \ \ \ \ \ \ \ \ \ \ \ \ \ 16.67\%} &  \includegraphics[width=0.33\textwidth]{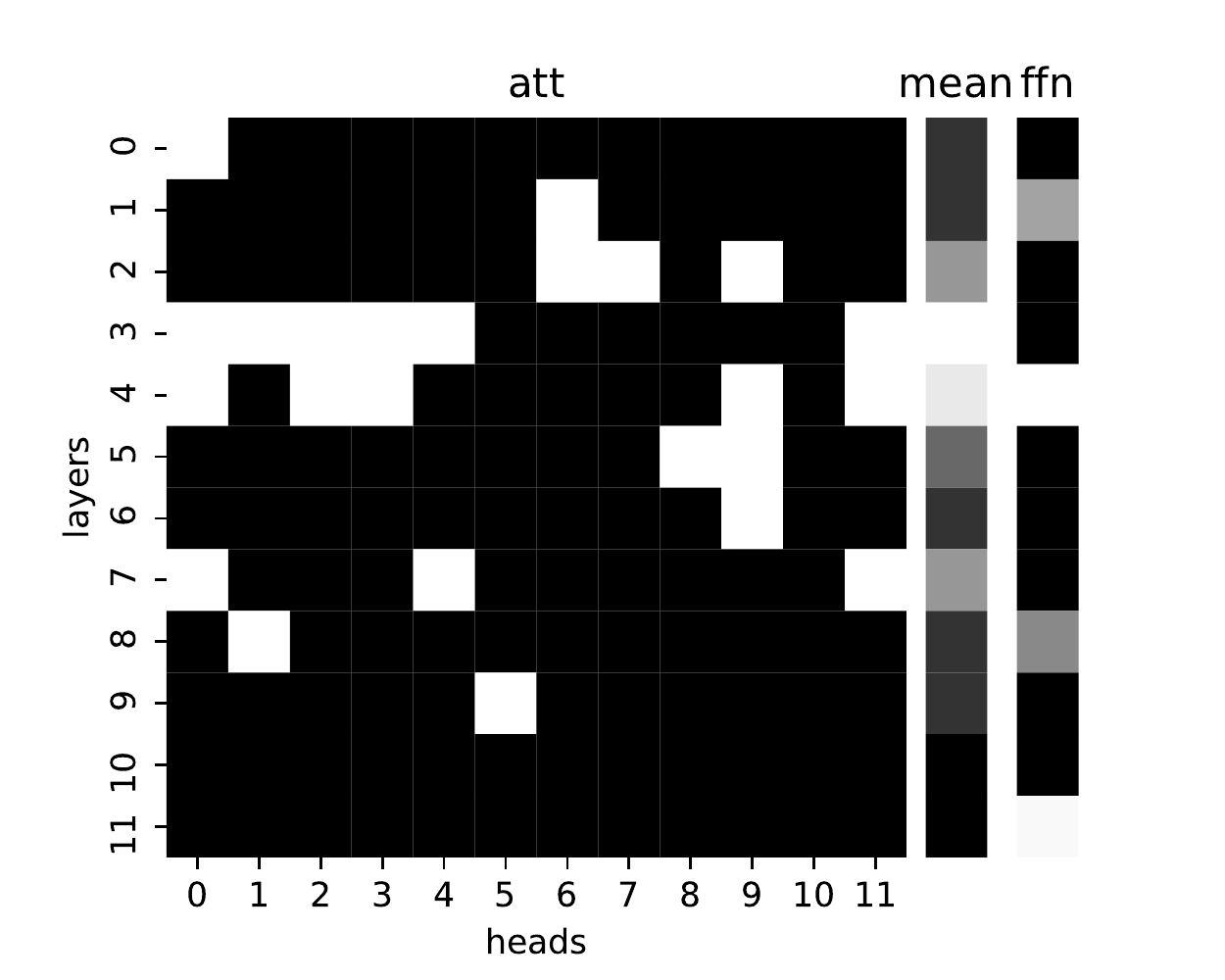}    & \includegraphics[width=0.33\textwidth]{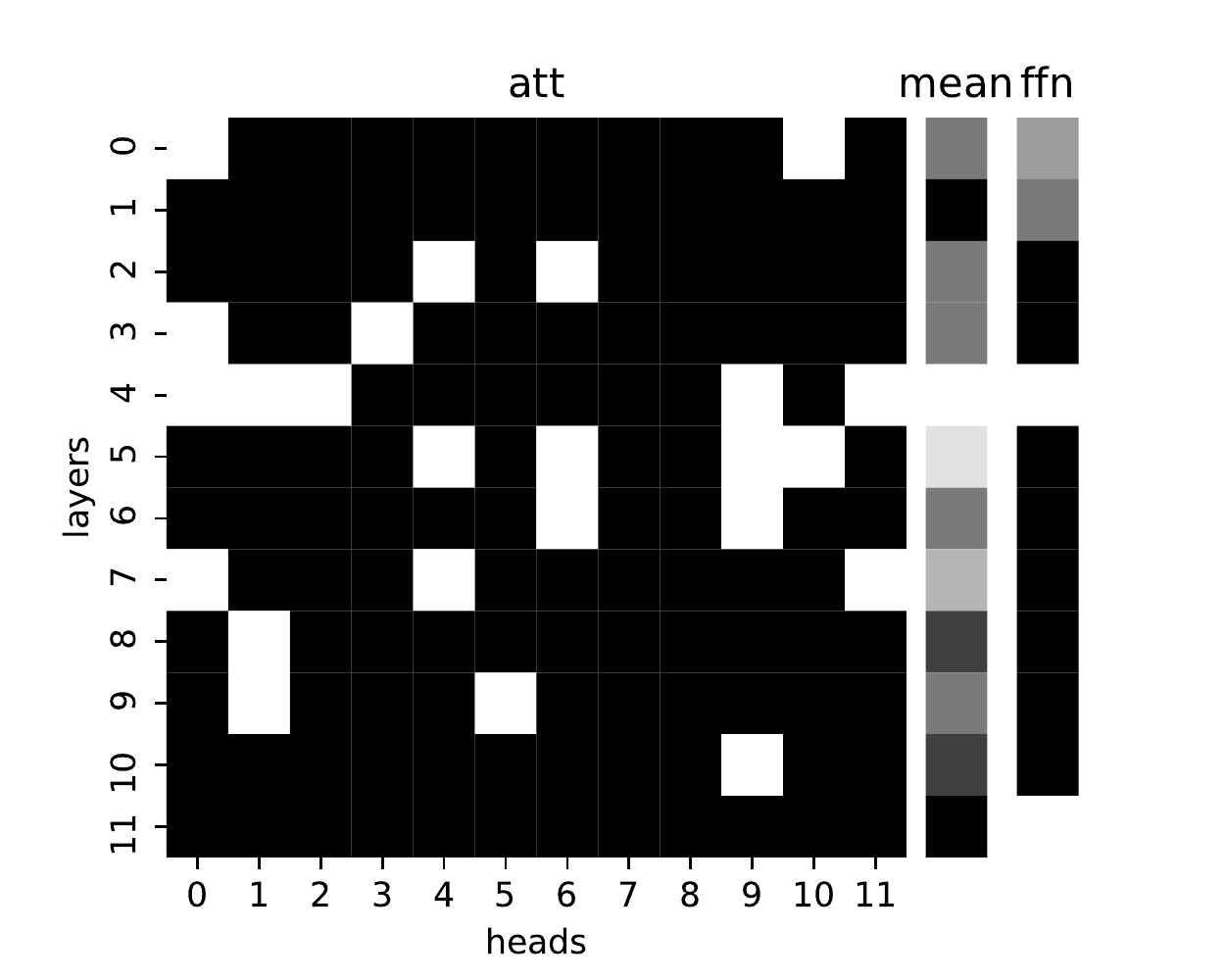} & \includegraphics[width=0.33\textwidth]{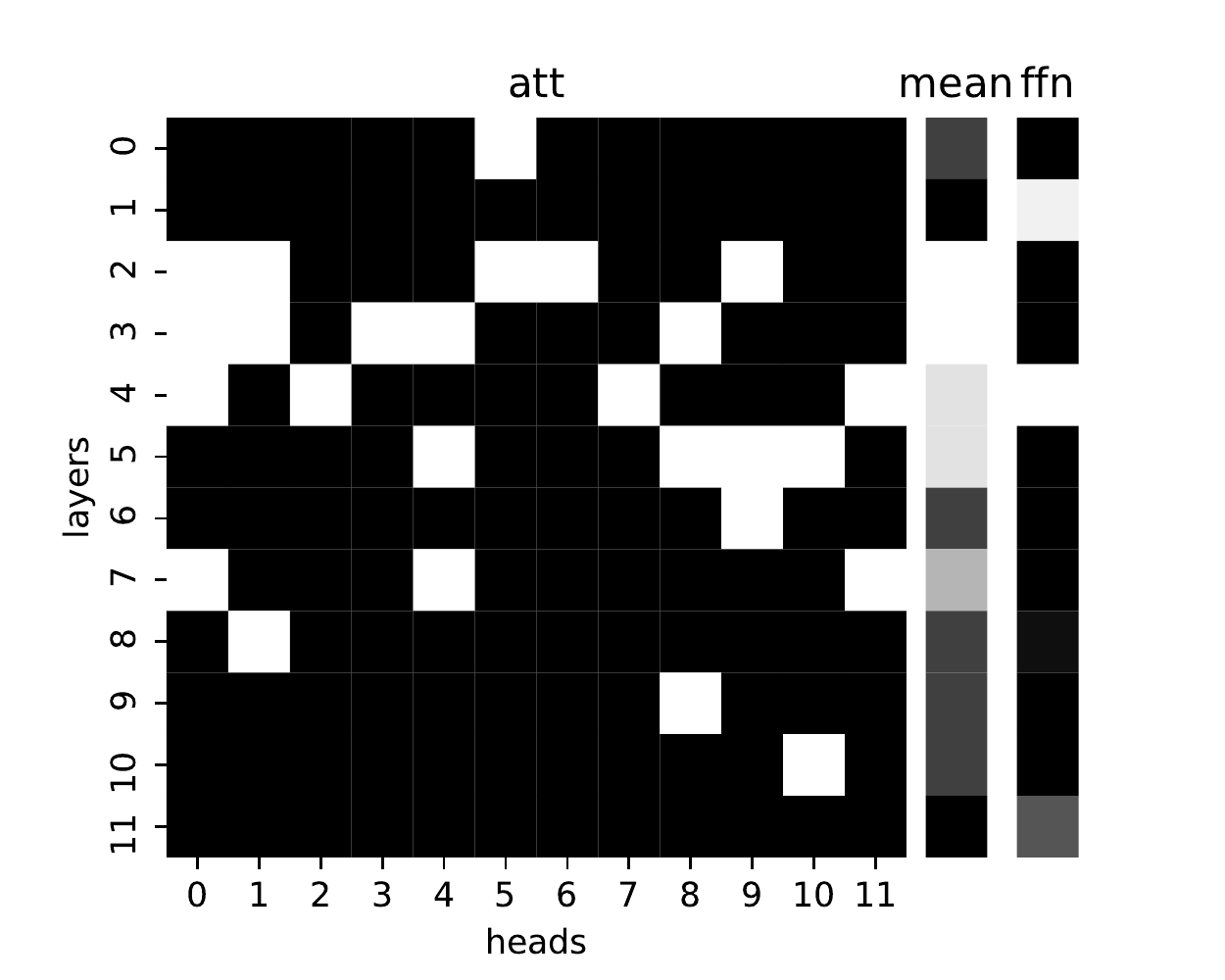} \\
     \rotatebox{90}{\ \ \ \ \ \ \ \ \ \ \ \ \ \ \ 33.33\%} &  \includegraphics[width=0.33\textwidth]{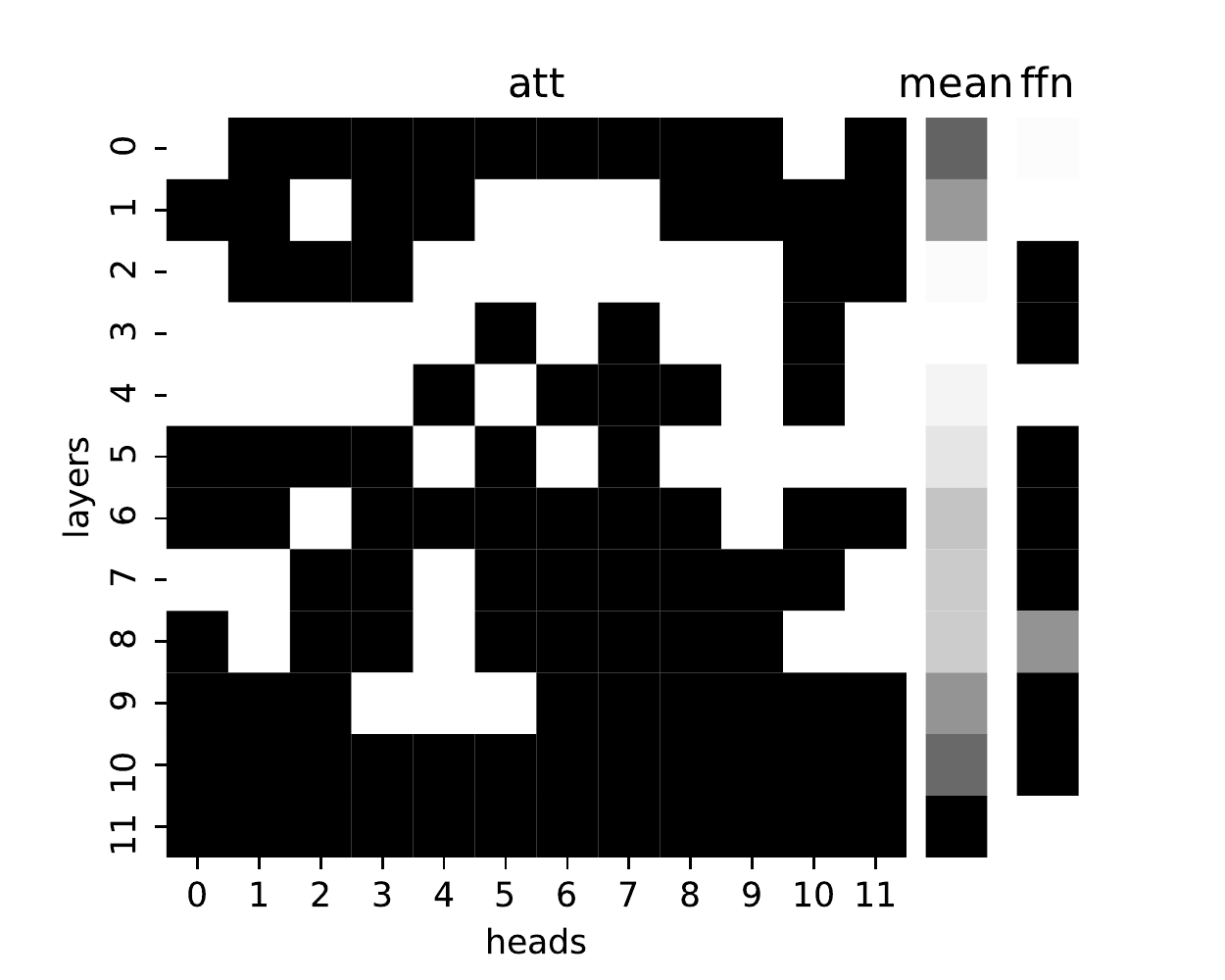}    & \includegraphics[width=0.33\textwidth]{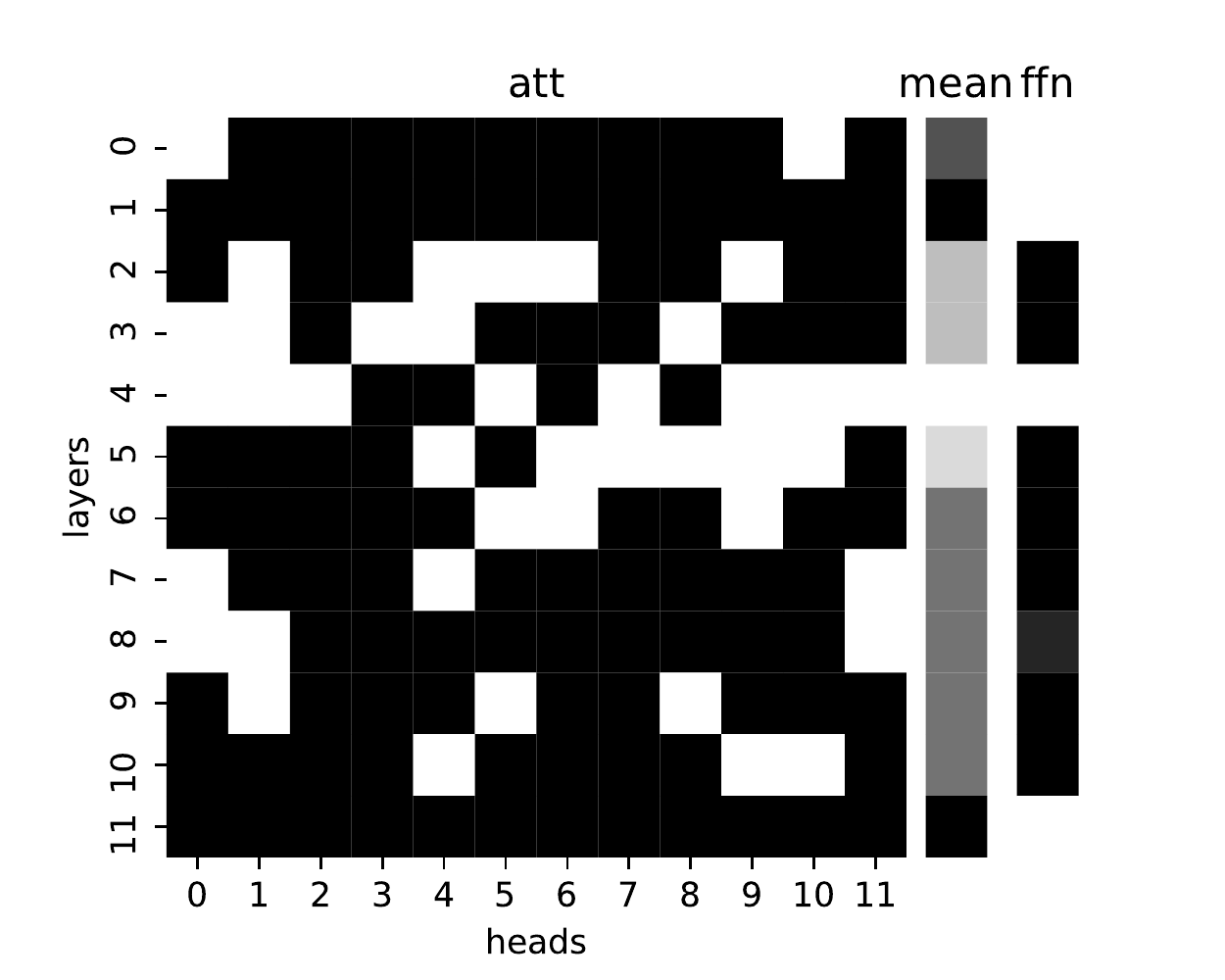} & \includegraphics[width=0.33\textwidth]{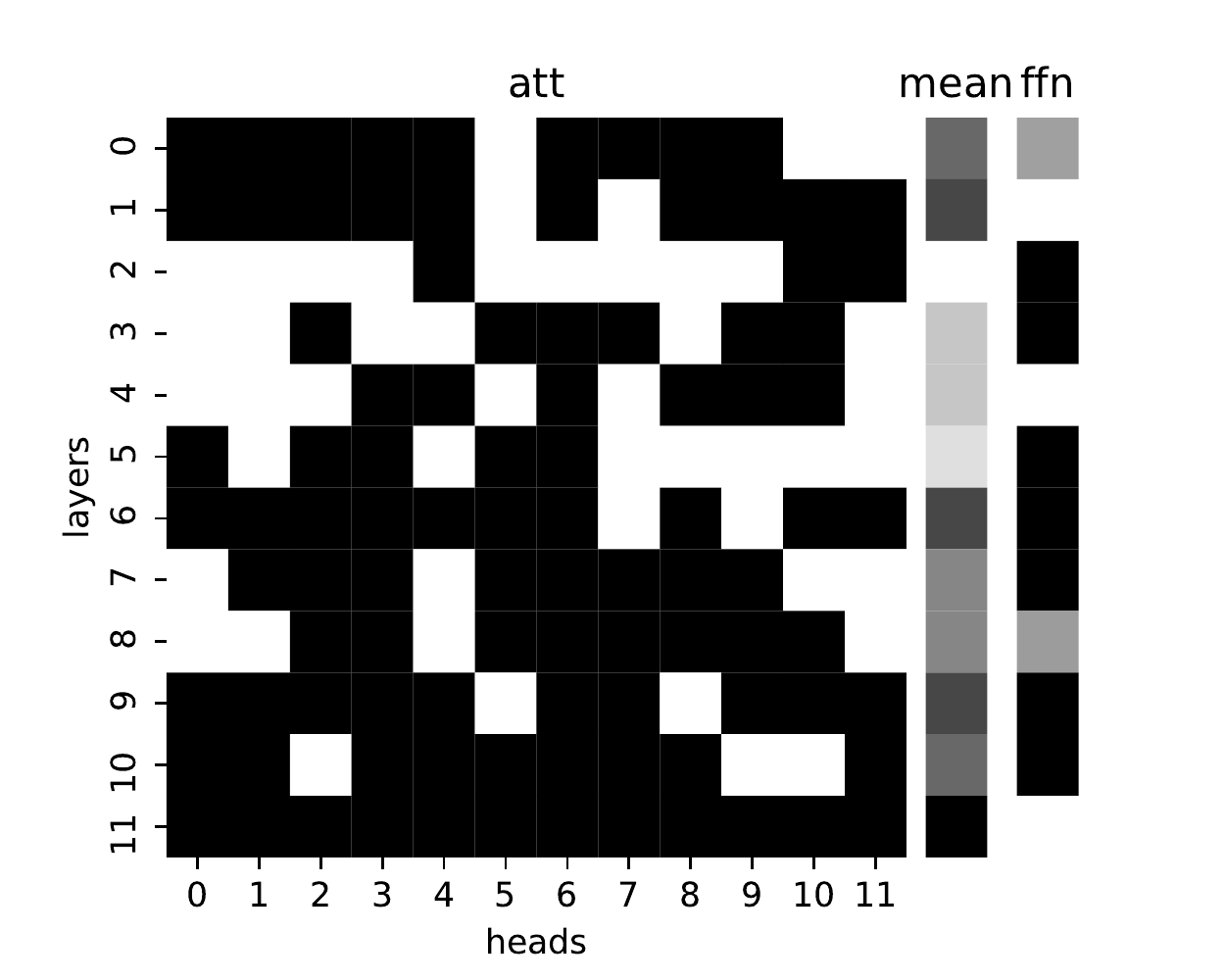} \\
     \rotatebox{90}{\ \ \ \ \ \ \ \ \ \ \ \ \ \ \ 50.00\%} &  \includegraphics[width=0.33\textwidth]{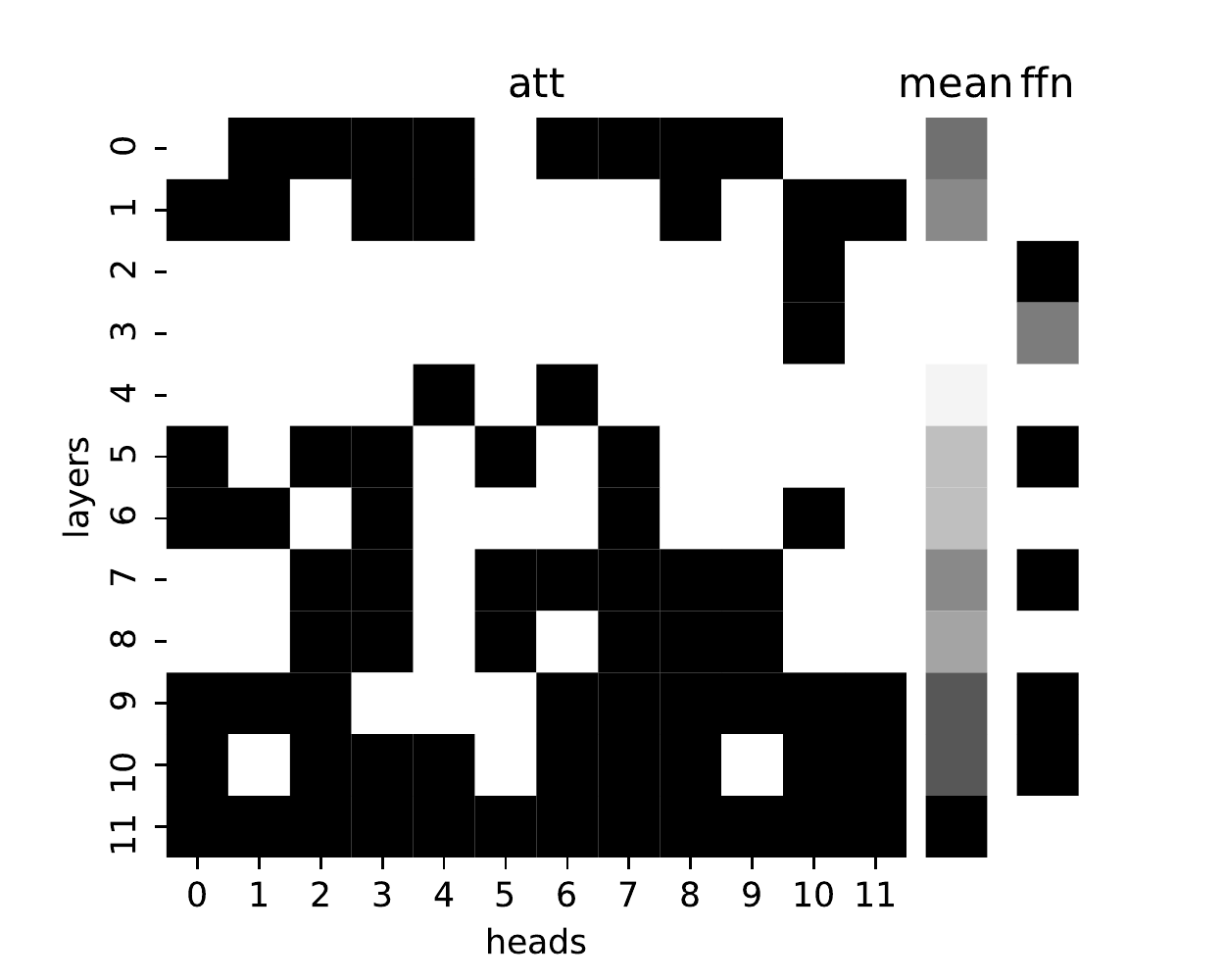}    & \includegraphics[width=0.33\textwidth]{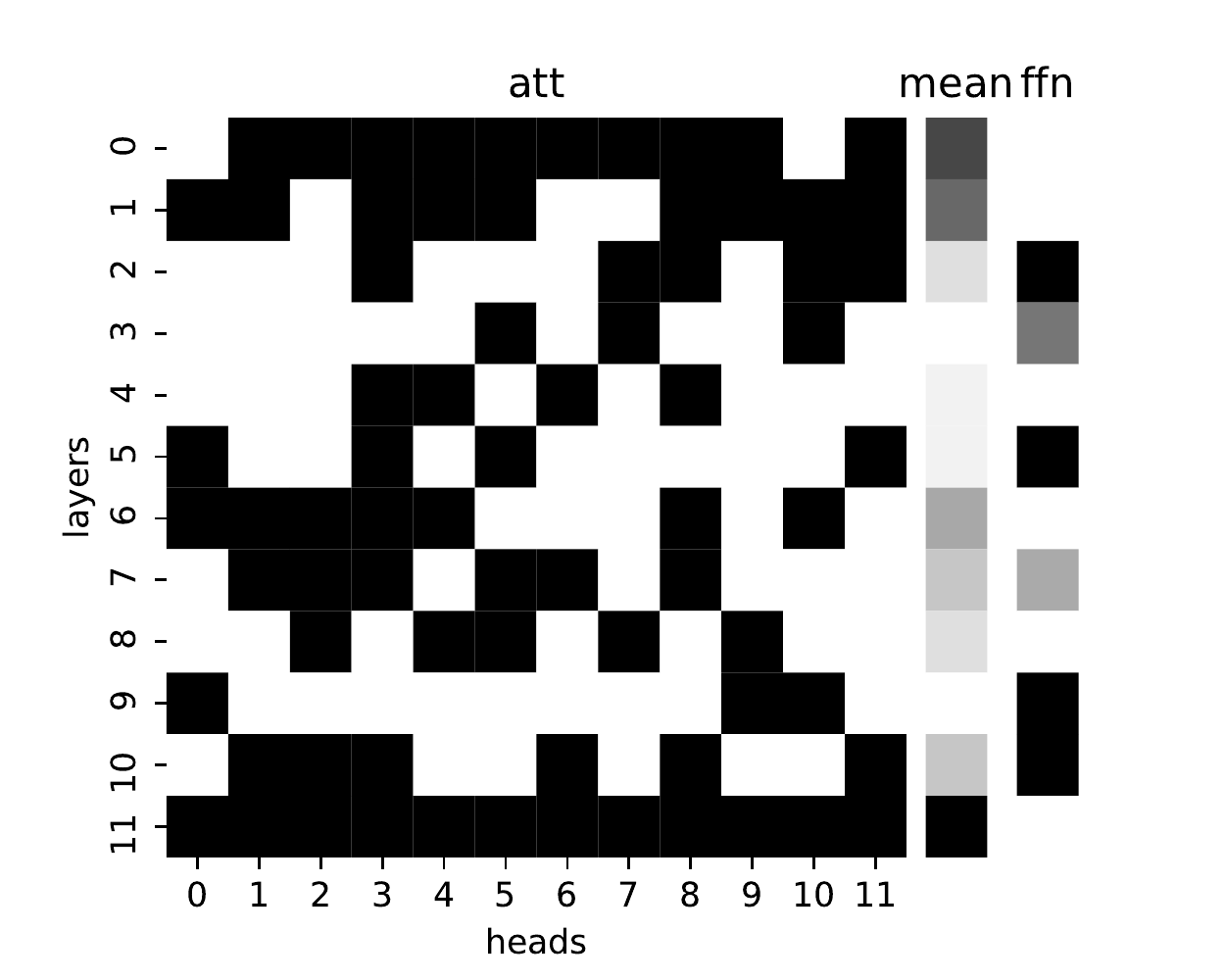} & \includegraphics[width=0.33\textwidth]{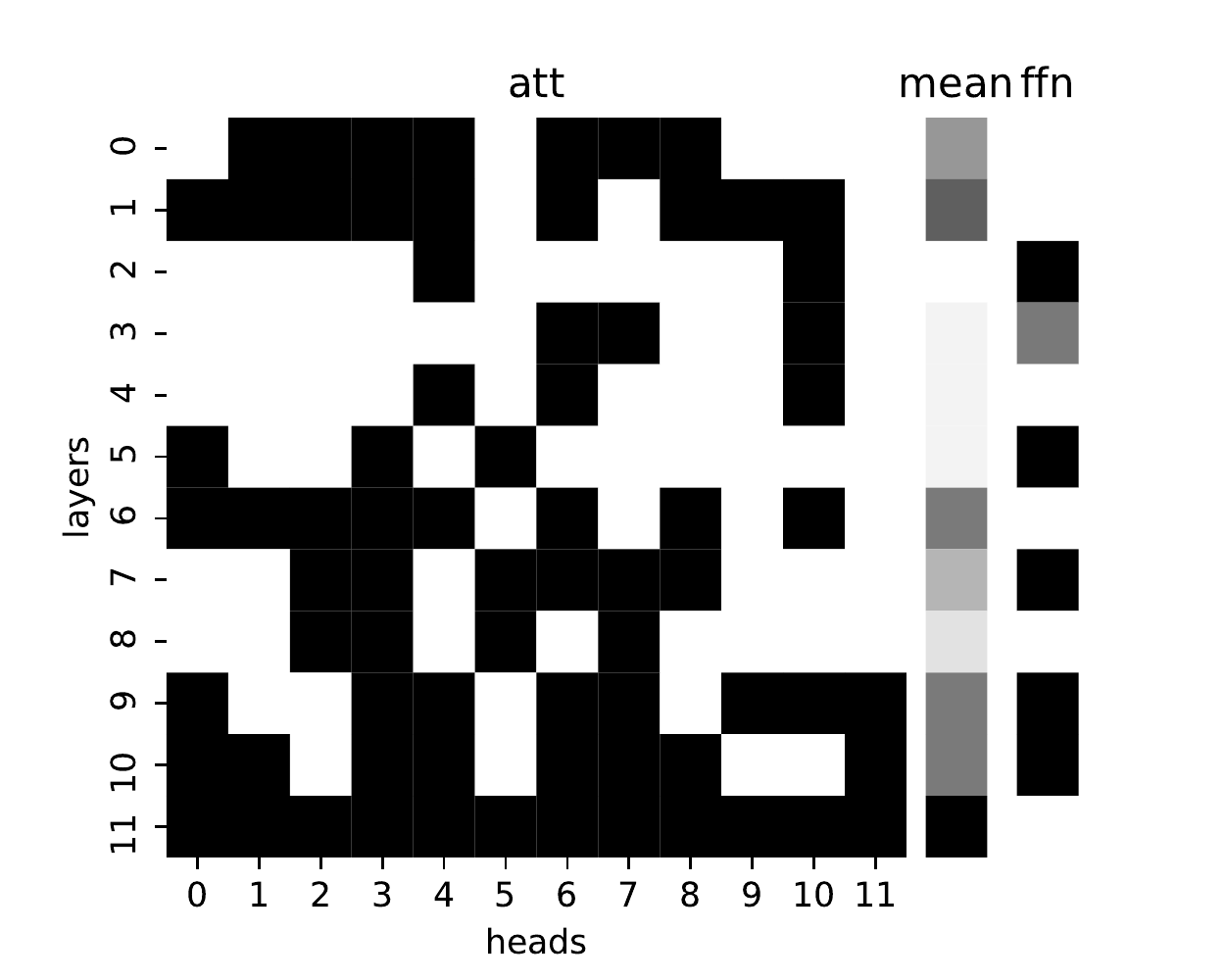}
    \end{tabular}
    \caption{
    Pruning map of attention heads (\texttt{att}) and FFN layers (\texttt{ffn}) during pruning on task SST-2, MRPC and MNLI dev sets. \texttt{mean} is the mean of attention heads of each layers. The left column shows the percentage of pruned parameters. Color white indicates that this attention head or FFN layer has been entirely pruned. For different tasks, the distribution of important attention heads is similar, but not identical, and certain FFN sub-layers might be more amenable to compression, even if the corresponding self-attention sub-unit is not.
    }
    \label{fig:heatmap}
\end{figure*}

\footnotetext[2]{TinyBERT \cite{jiao2019tinybert} utilizes data augmentation (DA), which makes it unfair to have a direct comparison. We only listed their results without DA.}
\footnotetext[3]{We do not list the results in \citet{guo2019reweighted, gordon2020compressing} since they show the results in scatter plots instead of the exact numbers, and similar to MvP \cite{sanh2020movement}, they are of unstructured pruning, which could not speed up the inference in practice.}

\subsection{Results}
We compare the pruning results on non-normalized and normalized BERT models on the 5 GLUE tasks, as shown in Figure \ref{fig:results}, which includes separate pruning for attention heads and FFN sub-layers and joint pruning of both modules. The results demonstrate the advantage of spectral normalization.

To put it all together, in Table \ref{tab:best-architecture}, we further show the simplest architecture we could get when allowed at most 1\% in terms of performance degradation. This also means that further slight pruning will have a noticeable impact on the final results. We find that spectral normalization can lead to a better trade-off between parameter size and performance. Specifically, for the same ideal performance shown in the last two rows in Table \ref{tab:best-architecture}, spectral normalized BERT could on average be pruned 12\% parameters more than the original BERT.

We also list other compression methods in the table for comparison while most of them are of knowledge distillation and have a pre-defined fixed size of the compressed model.
Our methods provide the flexibility to choose the best architecture and take advantage of finding the inflection point during pruning, while compared with other pruning methods we could practically speed up the inference time since we use a structured pruning.
\begin{figure}[tb]
    \centering
    \includegraphics[width=7.5cm]{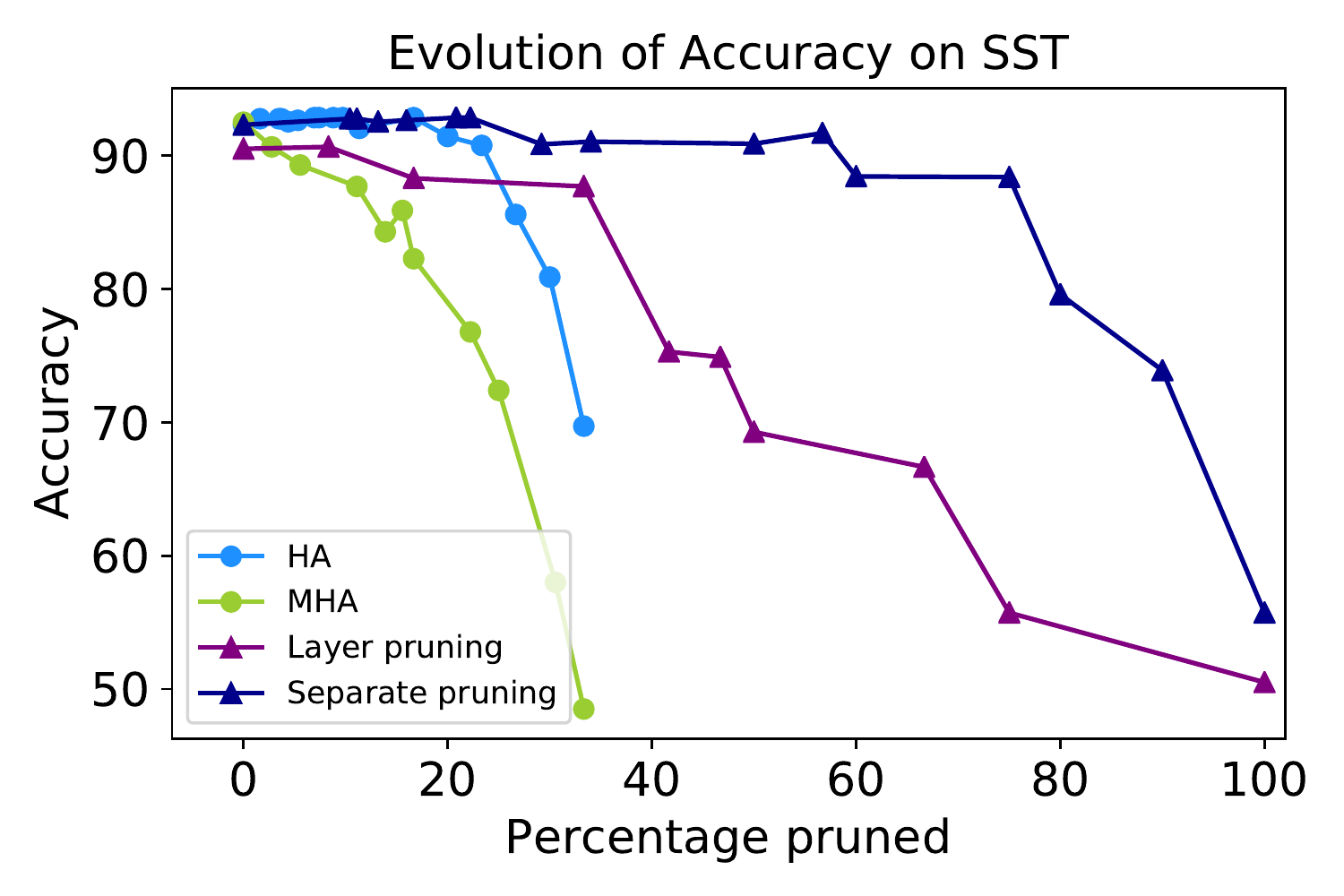}
    \caption{Evolution of accuracy on the dev set of SST-2 for (1) pruning the whole attention layers (HA); (2) pruning single attention heads (MHA); (3) pruning the whole layer (Layer pruning) and (4) pruning the attention and FFN sub-layers separately (Separate pruning).}
    \label{fig:abaltion}
\end{figure}

\subsection{Analysis}
In this section, we investigate the contribution of: (1) single attention head pruning, and (2) split pruning for attention heads and FFN sub-layers.

\paragraph{Single Head Attention Pruning}

Multi-head self-attention is a key component of Transformer, where each attention head potentially focuses on different parts of the inputs. 

The analysis of multi-head attention and its importance is challenging. 
Previous analysis of multi-head attention considered the average of attention weights over all heads at given position or focused only on the maximum attention weights \cite{voita2018context, tang2018analysis}, or explicitly takes into account the varying importance of different heads \cite{voita2019analyzing}. 
\citet{michel2019sixteen} has proved that attention heads can be removed without significantly impacting performance, but they mainly focus on machine translation and NLI.

To understand whether and to what extent attention heads play consistent and interpretable roles when trained on different downstream tasks, we pick one task from the sentiment analysis, paraphrasing and natural language inference respectively, plot the pruned attention heads during training and show the dynamic process in Figure \ref{fig:heatmap}. We can find that though for different tasks, the distributions of important attention heads are similar, as reflected in the mean of each attention layer, the exact pruned attention heads are actually different. This also indicates that splitting the pruning for attention heads could have more flexibility for the model to find an optimal architecture.

\paragraph{Separate Pruning for Attention and FFN}

Decoupling and then individually studying the self-attention and the FFN sub-layers is important for understanding the progressive improvement they provide.
As can be observed from Figure \ref{fig:results}, for most of the tasks, pruning FFN layers damages the performance more than the attention layers, indicating that the compression technique for the Transformer model tends to be more effective on the attention layers \cite{voita2018context, michel2019sixteen}, than the FFN layers \cite{ganesh2020compressing}.

In Figure \ref{fig:heatmap}, similar to attention heads, we further plot the pruning map of FFN layers. We find that, certain FFN sub-layers might be more amenable to compression, even if the corresponding self-attention sub-unit is not. For example, in all the tasks, the FFN layers near the ends of input or output are more likely to be pruned, while this does not hold for the corresponding self-attention layers.

Finally, we compare between single head attention pruning and separate pruning for attention/FFN, and show the evolution of performance for single head attention pruning (HA), multi-head attention pruning (MHA), separate attention/FFN pruning, and whole layer pruning respectively in Figure \ref{fig:abaltion}. We find that MHA and separate pruning perform much better than HA and layer pruning.

\section{Conclusion}
In this work, we propose a structured pruning method for compressing Transformer models, which prunes redundant mappings via spectral-normalized identity priors (SNIP).
We achieve effective pruning results on BERT fine-tuning while maintaining comparable performance.
Our work shows the importance of the mathematical properties of the Transformer model (specifically, the Lipschitz condition) on the effectiveness of pruning.

Additionally, we quantify task-specific trade-offs between model complexity and task performance, as well as the progressive improvement provided by Multi-head Attention (MHA) and Feedforward Networks (FFN).
Our results show that applying pruning at the level of mappings instead of individual weights allows for better model compression, when combined with the appropriate regularization. 
This suggests that developing more global pruning strategies may be a fruitful avenue for future research.

In the future, we plan to apply a similar approach to further reduce the width of Transformer layers, i.e., the hidden dimension, to achieve an even higher compression ratio. We are also interested in jointly using the proposed approach with other compression methods.

\section*{Acknowledgement}
We thank Atish Agarwala, Xikun Zhang, and anonymous reviewers for helpful feedback.
\bibliography{ref}

\begin{thebibliography}{53}
\expandafter\ifx\csname natexlab\endcsname\relax\def\natexlab#1{#1}\fi

\bibitem[{Ba and Caruana(2014)}]{ba2014deep}
Jimmy Ba and Rich Caruana. 2014.
\newblock Do deep nets really need to be deep?
\newblock In \emph{Advances in neural information processing systems}, pages
  2654--2662.

\bibitem[{Ba et~al.(2016)Ba, Kiros, and Hinton}]{ba2016layer}
Jimmy~Lei Ba, Jamie~Ryan Kiros, and Geoffrey~E Hinton. 2016.
\newblock Layer normalization.
\newblock \emph{arXiv preprint arXiv:1607.06450}.

\bibitem[{Behrmann et~al.(2019)Behrmann, Grathwohl, Chen, Duvenaud, and
  Jacobsen}]{behrmann_invertible_2019}
Jens Behrmann, Will Grathwohl, Ricky T.~Q. Chen, David Duvenaud, and
  Joern-Henrik Jacobsen. 2019.
\newblock \href {http://proceedings.mlr.press/v97/behrmann19a.html} {Invertible
  {Residual} {Networks}}.
\newblock In \emph{International {Conference} on {Machine} {Learning}}, pages
  573--582.
\newblock ISSN: 1938-7228 Section: Machine Learning.

\bibitem[{Chen et~al.(2020)Chen, Li, Qiu, Wang, Li, Ding, Deng, Huang, Lin, and
  Zhou}]{chen2020adabert}
Daoyuan Chen, Yaliang Li, Minghui Qiu, Zhen Wang, Bofang Li, Bolin Ding, Hongbo
  Deng, Jun Huang, Wei Lin, and Jingren Zhou. 2020.
\newblock Adabert: Task-adaptive bert compression with differentiable neural
  architecture search.
\newblock \emph{arXiv preprint arXiv:2001.04246}.

\bibitem[{Chen et~al.(2018)Chen, Zhang, Zhang, and Zhao}]{chen2018quora}
Zihan Chen, Hongbo Zhang, Xiaoji Zhang, and Leqi Zhao. 2018.
\newblock Quora question pairs.

\bibitem[{Cisse et~al.(2017)Cisse, Bojanowski, Grave, Dauphin, and
  Usunier}]{cisse2017parseval}
Moustapha Cisse, Piotr Bojanowski, Edouard Grave, Yann Dauphin, and Nicolas
  Usunier. 2017.
\newblock Parseval networks: Improving robustness to adversarial examples.
\newblock \emph{arXiv preprint arXiv:1704.08847}.

\bibitem[{Clark et~al.(2020)Clark, Luong, Le, and Manning}]{clark2020electra}
Kevin Clark, Minh-Thang Luong, Quoc~V Le, and Christopher~D Manning. 2020.
\newblock Electra: Pre-training text encoders as discriminators rather than
  generators.
\newblock \emph{arXiv preprint arXiv:2003.10555}.

\bibitem[{Devlin et~al.(2018)Devlin, Chang, Lee, and
  Toutanova}]{devlin2018bert}
Jacob Devlin, Ming-Wei Chang, Kenton Lee, and Kristina Toutanova. 2018.
\newblock Bert: Pre-training of deep bidirectional transformers for language
  understanding.
\newblock \emph{arXiv preprint arXiv:1810.04805}.

\bibitem[{Dolan and Brockett(2005)}]{dolan2005automatically}
William~B Dolan and Chris Brockett. 2005.
\newblock Automatically constructing a corpus of sentential paraphrases.
\newblock In \emph{Proceedings of the Third International Workshop on
  Paraphrasing (IWP2005)}.

\bibitem[{Fan et~al.(2020)Fan, Grave, and Joulin}]{fan_reducing_2020}
Angela Fan, Edouard Grave, and Armand Joulin. 2020.
\newblock \href {https://openreview.net/forum?id=SylO2yStDr} {Reducing
  {Transformer} {Depth} on {Demand} with {Structured} {Dropout}}.
\newblock In \emph{International {Conference} on {Learning} {Representations}}.

\bibitem[{Farnia et~al.(2018)Farnia, Zhang, and
  Tse}]{farnia_generalizable_2018}
Farzan Farnia, Jesse Zhang, and David Tse. 2018.
\newblock Generalizable adversarial training via spectral normalization.
\newblock In \emph{International Conference on Learning Representations}.

\bibitem[{Ganesh et~al.(2020)Ganesh, Chen, Lou, Khan, Yang, Chen, Winslett,
  Sajjad, and Nakov}]{ganesh2020compressing}
Prakhar Ganesh, Yao Chen, Xin Lou, Mohammad~Ali Khan, Yin Yang, Deming Chen,
  Marianne Winslett, Hassan Sajjad, and Preslav Nakov. 2020.
\newblock Compressing large-scale transformer-based models: A case study on
  bert.
\newblock \emph{arXiv preprint arXiv:2002.11985}.

\bibitem[{Gordon et~al.(2020)Gordon, Duh, and Andrews}]{gordon2020compressing}
Mitchell~A Gordon, Kevin Duh, and Nicholas Andrews. 2020.
\newblock Compressing bert: Studying the effects of weight pruning on transfer
  learning.
\newblock \emph{arXiv preprint arXiv:2002.08307}.

\bibitem[{Guo et~al.(2019)Guo, Liu, Mungall, Lin, and Wang}]{guo2019reweighted}
Fu-Ming Guo, Sijia Liu, Finlay~S Mungall, Xue Lin, and Yanzhi Wang. 2019.
\newblock Reweighted proximal pruning for large-scale language representation.
\newblock \emph{arXiv preprint arXiv:1909.12486}.

\bibitem[{Han et~al.(2015)Han, Pool, Tran, and Dally}]{han2015learning}
Song Han, Jeff Pool, John Tran, and William Dally. 2015.
\newblock Learning both weights and connections for efficient neural network.
\newblock In \emph{Advances in neural information processing systems}, pages
  1135--1143.

\bibitem[{He et~al.(2016{\natexlab{a}})He, Zhang, Ren, and Sun}]{he2016deep}
Kaiming He, Xiangyu Zhang, Shaoqing Ren, and Jian Sun. 2016{\natexlab{a}}.
\newblock Deep residual learning for image recognition.
\newblock In \emph{Proceedings of the IEEE conference on computer vision and
  pattern recognition}, pages 770--778.

\bibitem[{He et~al.(2016{\natexlab{b}})He, Zhang, Ren, and
  Sun}]{he2016identity}
Kaiming He, Xiangyu Zhang, Shaoqing Ren, and Jian Sun. 2016{\natexlab{b}}.
\newblock Identity mappings in deep residual networks.
\newblock In \emph{European conference on computer vision}, pages 630--645.
  Springer.

\bibitem[{Hinton et~al.(2015)Hinton, Vinyals, and Dean}]{hinton2015distilling}
Geoffrey Hinton, Oriol Vinyals, and Jeff Dean. 2015.
\newblock Distilling the knowledge in a neural network.
\newblock \emph{arXiv preprint arXiv:1503.02531}.

\bibitem[{Iandola et~al.(2016)Iandola, Han, Moskewicz, Ashraf, Dally, and
  Keutzer}]{iandola2016squeezenet}
Forrest~N Iandola, Song Han, Matthew~W Moskewicz, Khalid Ashraf, William~J
  Dally, and Kurt Keutzer. 2016.
\newblock Squeezenet: Alexnet-level accuracy with 50x fewer parameters and< 0.5
  mb model size.
\newblock \emph{arXiv preprint arXiv:1602.07360}.

\bibitem[{Jawahar et~al.(2019)Jawahar, Sagot, and Seddah}]{jawahar2019does}
Ganesh Jawahar, Beno{\^\i}t Sagot, and Djam{\'e} Seddah. 2019.
\newblock What does bert learn about the structure of language?
\newblock In \emph{Proceedings of the 57th Annual Meeting of the Association
  for Computational Linguistics}, pages 3651--3657.

\bibitem[{Jiao et~al.(2019)Jiao, Yin, Shang, Jiang, Chen, Li, Wang, and
  Liu}]{jiao2019tinybert}
Xiaoqi Jiao, Yichun Yin, Lifeng Shang, Xin Jiang, Xiao Chen, Linlin Li, Fang
  Wang, and Qun Liu. 2019.
\newblock Tinybert: Distilling bert for natural language understanding.
\newblock \emph{arXiv preprint arXiv:1909.10351}.

\bibitem[{Kovaleva et~al.(2019)Kovaleva, Romanov, Rogers, and
  Rumshisky}]{kovaleva2019revealing}
Olga Kovaleva, Alexey Romanov, Anna Rogers, and Anna Rumshisky. 2019.
\newblock Revealing the dark secrets of bert.
\newblock \emph{arXiv preprint arXiv:1908.08593}.

\bibitem[{Liu et~al.(2019{\natexlab{a}})Liu, Gardner, Belinkov, Peters, and
  Smith}]{liu2019linguistic}
Nelson~F Liu, Matt Gardner, Yonatan Belinkov, Matthew Peters, and Noah~A Smith.
  2019{\natexlab{a}}.
\newblock Linguistic knowledge and transferability of contextual
  representations.
\newblock \emph{arXiv preprint arXiv:1903.08855}.

\bibitem[{Liu et~al.(2019{\natexlab{b}})Liu, Ott, Goyal, Du, Joshi, Chen, Levy,
  Lewis, Zettlemoyer, and Stoyanov}]{liu2019roberta}
Yinhan Liu, Myle Ott, Naman Goyal, Jingfei Du, Mandar Joshi, Danqi Chen, Omer
  Levy, Mike Lewis, Luke Zettlemoyer, and Veselin Stoyanov. 2019{\natexlab{b}}.
\newblock Roberta: A robustly optimized bert pretraining approach.
\newblock \emph{arXiv preprint arXiv:1907.11692}.

\bibitem[{Michel et~al.(2019)Michel, Levy, and Neubig}]{michel2019sixteen}
Paul Michel, Omer Levy, and Graham Neubig. 2019.
\newblock Are sixteen heads really better than one?
\newblock In \emph{Advances in Neural Information Processing Systems}, pages
  14014--14024.

\bibitem[{Miyato et~al.(2018)Miyato, Kataoka, Koyama, and
  Yoshida}]{miyato2018spectral}
Takeru Miyato, Toshiki Kataoka, Masanori Koyama, and Yuichi Yoshida. 2018.
\newblock Spectral normalization for generative adversarial networks.
\newblock \emph{arXiv preprint arXiv:1802.05957}.

\bibitem[{Neyshabur et~al.(2017)Neyshabur, Bhojanapalli, McAllester, and
  Srebro}]{neyshabur2017exploring}
Behnam Neyshabur, Srinadh Bhojanapalli, David McAllester, and Nati Srebro.
  2017.
\newblock Exploring generalization in deep learning.
\newblock In \emph{Advances in neural information processing systems}, pages
  5947--5956.

\bibitem[{Oberman and Calder(2018)}]{oberman2018lipschitz}
Adam~M Oberman and Jeff Calder. 2018.
\newblock Lipschitz regularized deep neural networks converge and generalize.
\newblock \emph{arXiv preprint arXiv:1808.09540}.

\bibitem[{Radford et~al.(2019)Radford, Wu, Child, Luan, Amodei, and
  Sutskever}]{radford2019language}
Alec Radford, Jeffrey Wu, Rewon Child, David Luan, Dario Amodei, and Ilya
  Sutskever. 2019.
\newblock Language models are unsupervised multitask learners.
\newblock \emph{OpenAI Blog}, 1(8):9.

\bibitem[{Sanh et~al.(2019)Sanh, Debut, Chaumond, and
  Wolf}]{sanh2019distilbert}
Victor Sanh, Lysandre Debut, Julien Chaumond, and Thomas Wolf. 2019.
\newblock Distilbert, a distilled version of bert: smaller, faster, cheaper and
  lighter.
\newblock \emph{arXiv preprint arXiv:1910.01108}.

\bibitem[{Sanh et~al.(2020)Sanh, Wolf, and Rush}]{sanh2020movement}
Victor Sanh, Thomas Wolf, and Alexander~M Rush. 2020.
\newblock Movement pruning: Adaptive sparsity by fine-tuning.
\newblock \emph{arXiv preprint arXiv:2005.07683}.

\bibitem[{Shen et~al.(2019)Shen, Dong, Ye, Ma, Yao, Gholami, Mahoney, and
  Keutzer}]{shen2019q}
Sheng Shen, Zhen Dong, Jiayu Ye, Linjian Ma, Zhewei Yao, Amir Gholami,
  Michael~W Mahoney, and Kurt Keutzer. 2019.
\newblock Q-bert: Hessian based ultra low precision quantization of bert.
\newblock \emph{arXiv preprint arXiv:1909.05840}.

\bibitem[{Socher et~al.(2013)Socher, Perelygin, Wu, Chuang, Manning, Ng, and
  Potts}]{socher2013recursive}
Richard Socher, Alex Perelygin, Jean Wu, Jason Chuang, Christopher~D Manning,
  Andrew~Y Ng, and Christopher Potts. 2013.
\newblock Recursive deep models for semantic compositionality over a sentiment
  treebank.
\newblock In \emph{Proceedings of the 2013 conference on empirical methods in
  natural language processing}, pages 1631--1642.

\bibitem[{Sokoli{\'c} et~al.(2017)Sokoli{\'c}, Giryes, Sapiro, and
  Rodrigues}]{sokolic2017robust}
Jure Sokoli{\'c}, Raja Giryes, Guillermo Sapiro, and Miguel~RD Rodrigues. 2017.
\newblock Robust large margin deep neural networks.
\newblock \emph{IEEE Transactions on Signal Processing}, 65(16):4265--4280.

\bibitem[{Srivastava et~al.(2015)Srivastava, Greff, and
  Schmidhuber}]{srivastava2015highway}
Rupesh~Kumar Srivastava, Klaus Greff, and J{\"u}rgen Schmidhuber. 2015.
\newblock Highway networks.
\newblock \emph{arXiv preprint arXiv:1505.00387}.

\bibitem[{Sun et~al.(2019)Sun, Cheng, Gan, and Liu}]{sun2019patient}
Siqi Sun, Yu~Cheng, Zhe Gan, and Jingjing Liu. 2019.
\newblock Patient knowledge distillation for bert model compression.
\newblock \emph{arXiv preprint arXiv:1908.09355}.

\bibitem[{Sun et~al.(2020)Sun, Yu, Song, Liu, Yang, and
  Zhou}]{sun2020mobilebert}
Zhiqing Sun, Hongkun Yu, Xiaodan Song, Renjie Liu, Yiming Yang, and Denny Zhou.
  2020.
\newblock Mobilebert: a compact task-agnostic bert for resource-limited
  devices.
\newblock \emph{arXiv preprint arXiv:2004.02984}.

\bibitem[{Tang et~al.(2018)Tang, Sennrich, and Nivre}]{tang2018analysis}
Gongbo Tang, Rico Sennrich, and Joakim Nivre. 2018.
\newblock An analysis of attention mechanisms: The case of word sense
  disambiguation in neural machine translation.
\newblock \emph{arXiv preprint arXiv:1810.07595}.

\bibitem[{Tang et~al.(2019)Tang, Lu, Liu, Mou, Vechtomova, and
  Lin}]{tang2019distilling}
Raphael Tang, Yao Lu, Linqing Liu, Lili Mou, Olga Vechtomova, and Jimmy Lin.
  2019.
\newblock Distilling task-specific knowledge from bert into simple neural
  networks.
\newblock \emph{arXiv preprint arXiv:1903.12136}.

\bibitem[{Tenney et~al.(2019)Tenney, Das, and Pavlick}]{tenney2019bert}
Ian Tenney, Dipanjan Das, and Ellie Pavlick. 2019.
\newblock Bert rediscovers the classical nlp pipeline.
\newblock \emph{arXiv preprint arXiv:1905.05950}.

\bibitem[{Vaswani et~al.(2017)Vaswani, Shazeer, Parmar, Uszkoreit, Jones,
  Gomez, Kaiser, and Polosukhin}]{vaswani2017attention}
Ashish Vaswani, Noam Shazeer, Niki Parmar, Jakob Uszkoreit, Llion Jones,
  Aidan~N Gomez, {\L}ukasz Kaiser, and Illia Polosukhin. 2017.
\newblock Attention is all you need.
\newblock In \emph{Advances in neural information processing systems}, pages
  5998--6008.

\bibitem[{Veit and Belongie(2017)}]{veit2017convolutional}
Andreas Veit and Serge Belongie. 2017.
\newblock Convolutional networks with adaptive computation graphs.
\newblock \emph{arXiv preprint arXiv:1711.11503}, 2.

\bibitem[{Voita et~al.(2018)Voita, Serdyukov, Sennrich, and
  Titov}]{voita2018context}
Elena Voita, Pavel Serdyukov, Rico Sennrich, and Ivan Titov. 2018.
\newblock Context-aware neural machine translation learns anaphora resolution.
\newblock \emph{arXiv preprint arXiv:1805.10163}.

\bibitem[{Voita et~al.(2019)Voita, Talbot, Moiseev, Sennrich, and
  Titov}]{voita2019analyzing}
Elena Voita, David Talbot, Fedor Moiseev, Rico Sennrich, and Ivan Titov. 2019.
\newblock Analyzing multi-head self-attention: Specialized heads do the heavy
  lifting, the rest can be pruned.
\newblock \emph{arXiv preprint arXiv:1905.09418}.

\bibitem[{Wang et~al.(2018)Wang, Singh, Michael, Hill, Levy, and
  Bowman}]{wang2018glue}
Alex Wang, Amanpreet Singh, Julian Michael, Felix Hill, Omer Levy, and Samuel~R
  Bowman. 2018.
\newblock Glue: A multi-task benchmark and analysis platform for natural
  language understanding.
\newblock \emph{arXiv preprint arXiv:1804.07461}.

\bibitem[{Wang et~al.(2019)Wang, Wohlwend, and Lei}]{wang2019structured}
Ziheng Wang, Jeremy Wohlwend, and Tao Lei. 2019.
\newblock Structured pruning of large language models.
\newblock \emph{arXiv preprint arXiv:1910.04732}.

\bibitem[{Wen et~al.(2016)Wen, Wu, Wang, Chen, and Li}]{wen2016learning}
Wei Wen, Chunpeng Wu, Yandan Wang, Yiran Chen, and Hai Li. 2016.
\newblock Learning structured sparsity in deep neural networks.
\newblock In \emph{Advances in neural information processing systems}, pages
  2074--2082.

\bibitem[{Williams et~al.(2017)Williams, Nangia, and
  Bowman}]{williams2017broad}
Adina Williams, Nikita Nangia, and Samuel~R Bowman. 2017.
\newblock A broad-coverage challenge corpus for sentence understanding through
  inference.
\newblock \emph{arXiv preprint arXiv:1704.05426}.

\bibitem[{Yang et~al.(2019)Yang, Dai, Yang, Carbonell, Salakhutdinov, and
  Le}]{yang2019xlnet}
Zhilin Yang, Zihang Dai, Yiming Yang, Jaime Carbonell, Russ~R Salakhutdinov,
  and Quoc~V Le. 2019.
\newblock Xlnet: Generalized autoregressive pretraining for language
  understanding.
\newblock In \emph{Advances in neural information processing systems}, pages
  5754--5764.

\bibitem[{Yu et~al.(2018)Yu, Yu, and Ramalingam}]{yu2018learning}
Xin Yu, Zhiding Yu, and Srikumar Ramalingam. 2018.
\newblock Learning strict identity mappings in deep residual networks.
\newblock In \emph{Proceedings of the IEEE Conference on Computer Vision and
  Pattern Recognition}, pages 4432--4440.

\bibitem[{Zafrir et~al.(2019)Zafrir, Boudoukh, Izsak, and
  Wasserblat}]{zafrir2019q8bert}
Ofir Zafrir, Guy Boudoukh, Peter Izsak, and Moshe Wasserblat. 2019.
\newblock Q8bert: Quantized 8bit bert.
\newblock \emph{arXiv preprint arXiv:1910.06188}.

\bibitem[{Zhao et~al.(2018)Zhao, Kim, Zhang, Rush, and
  LeCun}]{zhao_adversarially_2018}
Junbo Zhao, Yoon Kim, Kelly Zhang, Alexander Rush, and Yann LeCun. 2018.
\newblock \href {http://proceedings.mlr.press/v80/zhao18b.html} {Adversarially
  {Regularized} {Autoencoders}}.
\newblock In \emph{International {Conference} on {Machine} {Learning}}, pages
  5902--5911.
\newblock ISSN: 1938-7228 Section: Machine Learning.

\bibitem[{Zhu et~al.(2015)Zhu, Kiros, Zemel, Salakhutdinov, Urtasun, Torralba,
  and Fidler}]{zhu2015aligning}
Yukun Zhu, Ryan Kiros, Rich Zemel, Ruslan Salakhutdinov, Raquel Urtasun,
  Antonio Torralba, and Sanja Fidler. 2015.
\newblock Aligning books and movies: Towards story-like visual explanations by
  watching movies and reading books.
\newblock In \emph{Proceedings of the IEEE international conference on computer
  vision}, pages 19--27.

\end{thebibliography}
\bibliographystyle{acl_natbib}

\appendix

\section{Appendices}
\label{sec:appendix}

\begin{figure*}[t]
    \begin{minipage}[t]{0.33\textwidth}
    \centering
        \includegraphics[width=1.0\textwidth]{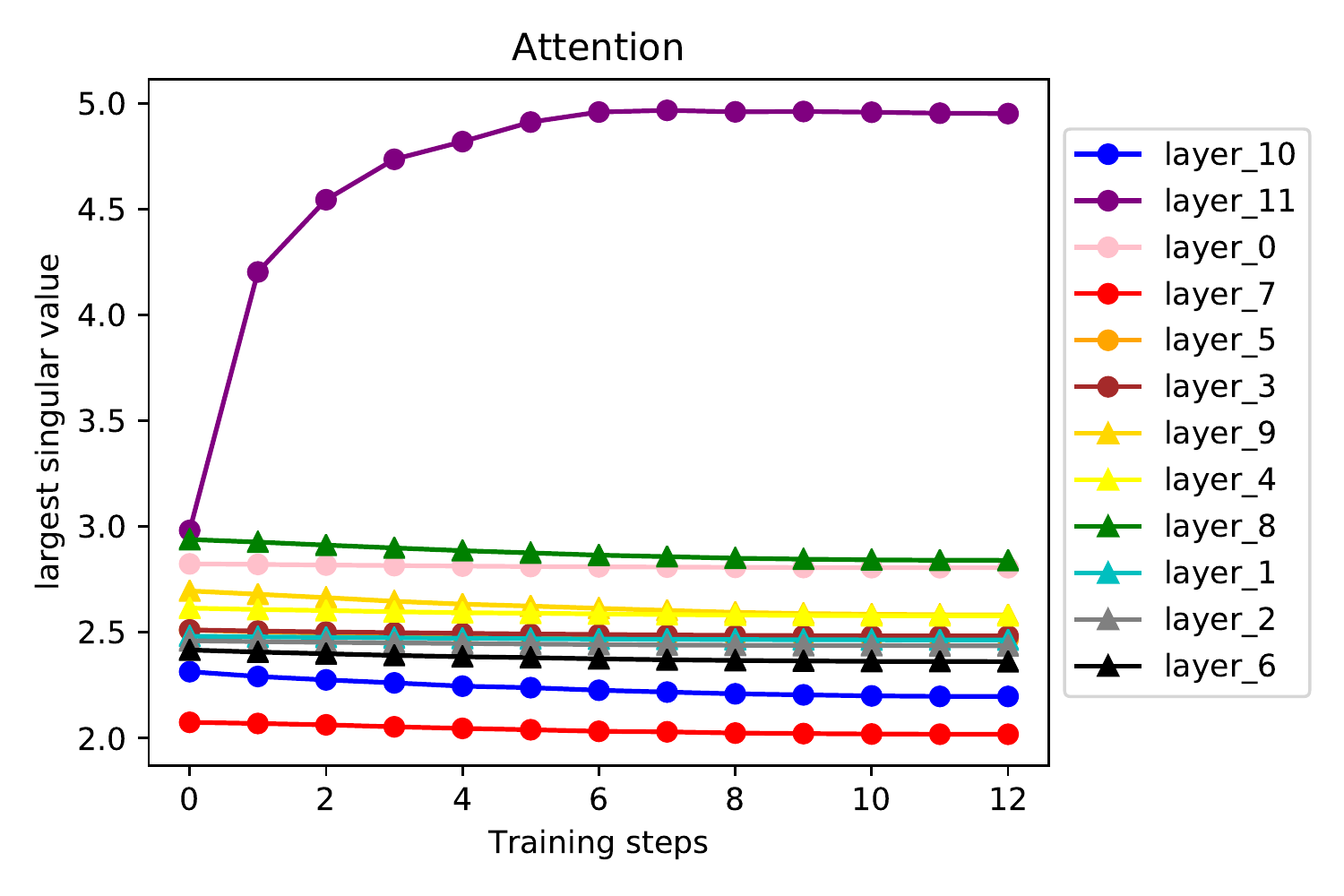}
    \end{minipage}%
    \begin{minipage}[t]{0.33\textwidth}
    \centering
        \includegraphics[width=1.0\textwidth]{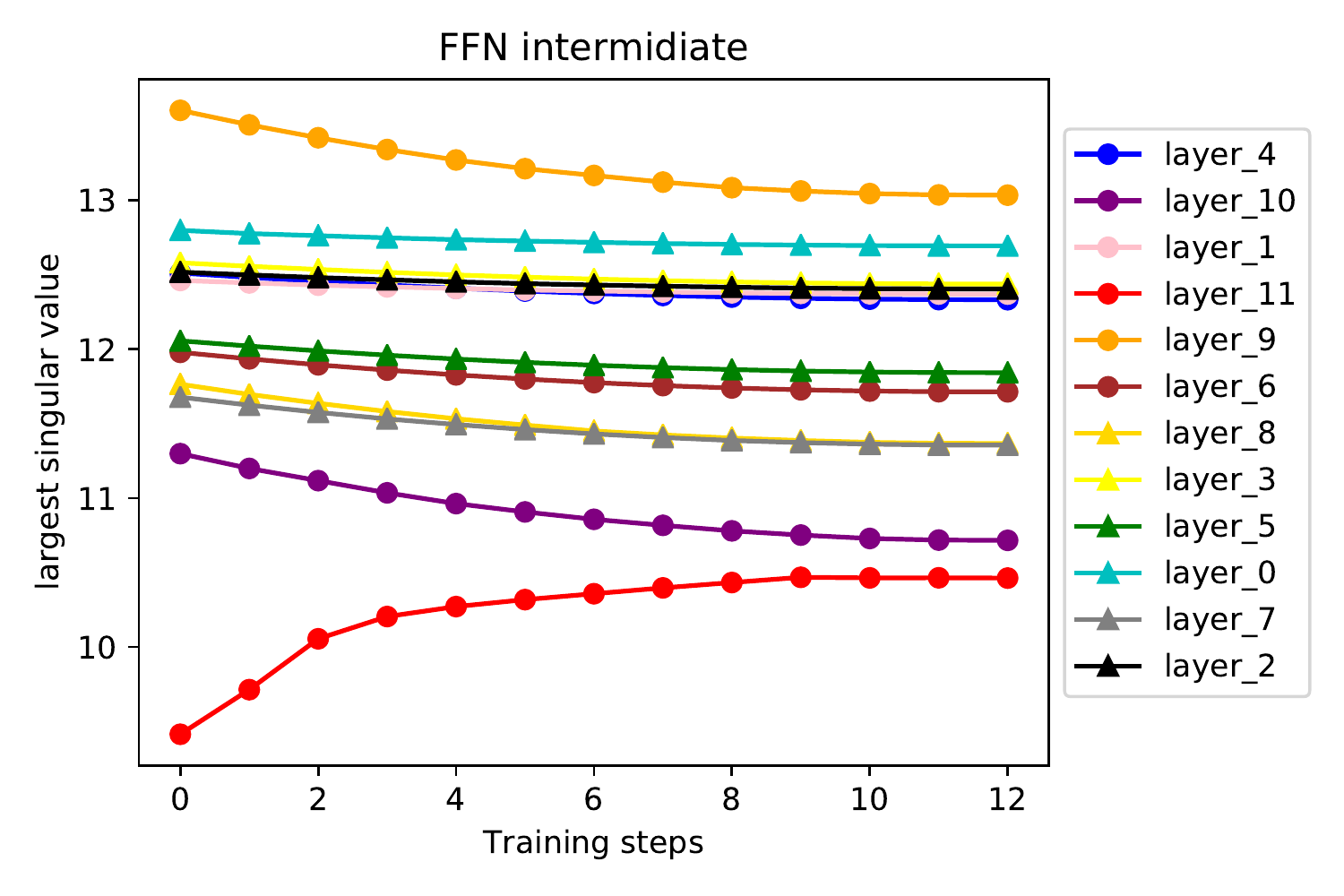}
    \end{minipage}
    \begin{minipage}[t]{0.33\textwidth}
    \centering
        \includegraphics[width=1.0\textwidth]{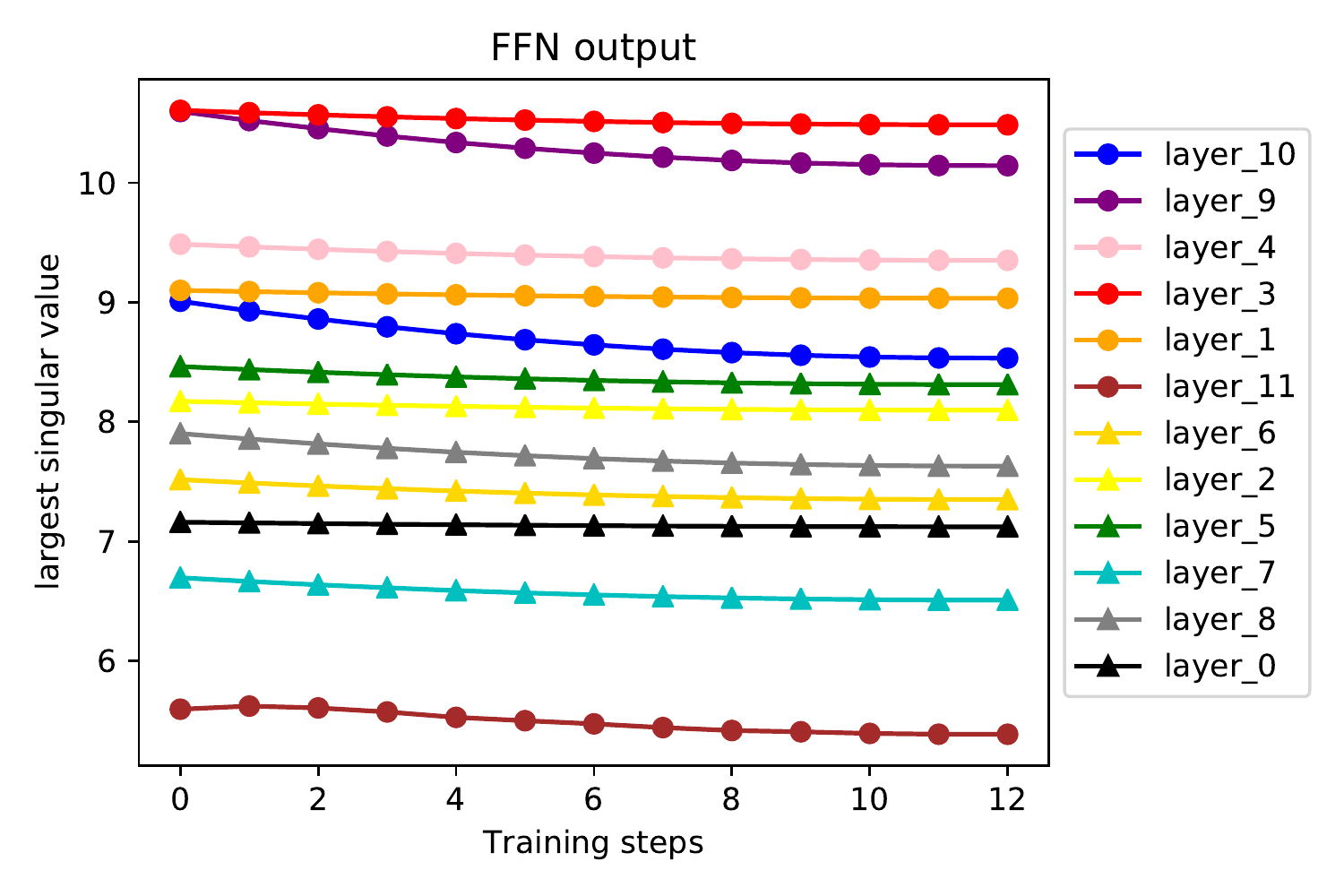}
    \end{minipage}
    \caption{The largest singular value of attention output weights, and FFN intermediate and output weights for the original BERT$_\textrm{BASE}$ during the fine-tuning on the SST-2 task.}
    \label{fig:largest-singular-value}
\end{figure*}

\subsection{GLUE Dataset}
We provide a brief description of the 5 tasks in our experiments from the GLUE benchmarks \citep{wang2018glue}.

\paragraph{SST-2} The Stanford Sentiment Treebank \cite{socher2013recursive} consists of sentences from movie reviews and human annotations of their sentiment. The task is to predict the sentiment of a given sentence (positive/negative). The performance is evaluated by the accuracy.

\paragraph{QQP} The Quora Question Pairs dataset \cite{chen2018quora} is a collection of question pairs from the community question-answering website Quora. The task is to determine whether a pair of questions are semantically equivalent. The performance is evaluated by the accuracy.

\begin{table}[tb]
    \centering
    \scalebox{0.75}{
    \begin{tabular}{l|l|l}
    \hline
    \textbf{Hyperparameter}     &  \textbf{Pre-training} & \textbf{Fine-tuning}\\\hline
     Number of layers    & 12  & 12\\
     Hidden size & 768 & 768\\
     FFN inner hidden size & 3072 & 3072 \\
     Attention heads & 12 & 12\\
     Attention head size & 64 & 64\\
     Embedding size & 768 & 768\\
     Mask percent & 15 & -\\
     Learning rate decay & linear & linear\\
     Layerwise LR decay & - & 0.8 \\
     Warmup steps & 10000 & -\\
     Warmup fraction & - & 0.1 \\
     Learning Rate & 2e-4 & 1e-4\\
     Adam $\epsilon$ & 1e-6 & 1e-6\\
     Adam $\beta_1$ & 0.9 & 0.9\\
     Adam & 0.999 & 0.999\\
     Attention dropout & 0.1 & 0.1\\
     Dropout & 0.1 & 0.1\\
     Weight decay & 0.01 & 0.01\\
     $\textit{l}_1$ regularization factor & - & 0.01 \\
     Batch size & 2048 & 32\\
     Train steps & 1M & -\\
     Train epochs & -  & 3.0/iter for MRPC,  \\
     & & 1.0 for others \\\hline
     \end{tabular}}
    \caption{Pre-training and Fine-tuning hyperparameters}
    \label{tab:hp-pre-train}
\end{table}

\paragraph{MRPC} The Microsoft Research Paraphrase Corpus \cite{dolan2005automatically} is a corpus of sentence pairs automatically extracted from online news sources, with human annotations for whether the sentences in the pair are semantically equivalent, and the task is to predict the equivalence. The performance is evaluated by both the F1 score.

\paragraph{QNLI} The Question-answering NLI dataset \cite{chen2018quora} is converted from the Stanford Question Answering Dataset (SQuAD) to a classification task. The performance is evaluated by the accuracy.

\paragraph{MNLI} The Multi-Genre Natural Language Inference Corpus \cite{williams2017broad} is a crowdsourced collection of sentence pairs with textual entailment annotations. Given a premise sentence and a hypothesis sentence, the task is to predict whether the premise entails the hypothesis (\textit{entailment}), contradicts the hypothesis (\textit{contradiction}), or neither (\textit{neutral}). The performance is evaluated by the test accuracy on both matched (in-domain) and mismatched (cross-domain) sections of the test data.

\subsection{Experiment Settings}
The full set of hyperparameters for pre-training and fine-tuning are listed in Table \ref{tab:hp-pre-train}.

\subsection{Spectral Norm of Weights during Training}
We show the largest singular values of the weight metrics in the original BERT model during fine-tuning on the task SST-2 in Figure \ref{fig:largest-singular-value}. As can be seen from the figure, the norm of the weights without spectral normalization is obviously out of control.


\end{document}